\pdfoutput=1

\documentclass[11pt]{article}

\usepackage[preprint]{acl}

\usepackage{times}
\usepackage{latexsym}
\usepackage[normalem]{ulem}
\useunder{\uline}{\ul}{}
\usepackage[T1]{fontenc}

\usepackage[utf8]{inputenc}

\usepackage{microtype}

\usepackage{inconsolata}

\usepackage{graphicx}

\usepackage{multirow}
\usepackage{amssymb}
\usepackage{pifont}
\usepackage{amsmath}
\usepackage{color}
\usepackage[normalem]{ulem}
\useunder{\uline}{\ul}{}

\newcommand{\cmark}{\ding{51}}%
\newcommand{\xmark}{\ding{55}}%
\title{ReviewAgents: Bridging the Gap Between Human and AI-Generated Paper Reviews}

\author{
 \textbf{Xian Gao\textsuperscript{1}},
 \textbf{Jiacheng Ruan\textsuperscript{1}},
 \textbf{Zongyun Zhang\textsuperscript{1}},
 \textbf{Jingsheng Gao\textsuperscript{1}},
 \textbf{Ting Liu\textsuperscript{1}},
 \textbf{Yuzhuo Fu\textsuperscript{1}}
\\
 \textsuperscript{1}Shanghai Jiao Tong University
}

\begin{document}
\maketitle

\begin{abstract}

Academic paper review is a critical yet time-consuming task within the research community. With the increasing volume of academic publications, automating the review process has become a significant challenge. The primary issue lies in generating comprehensive, accurate, and reasoning-consistent review comments that align with human reviewers' judgments. In this paper, we address this challenge by proposing \textbf{ReviewAgents}, a framework that leverages large language models (LLMs) to generate academic paper reviews. We first introduce a novel dataset, \textbf{Review-CoT}, consisting of 142k review comments, designed for training LLM agents. This dataset emulates the structured reasoning process of human reviewers—summarizing the paper, referencing relevant works, identifying strengths and weaknesses, and generating a review conclusion. Building upon this, we train LLM reviewer agents capable of structured reasoning using a relevant-paper-aware training method. Furthermore, we construct ReviewAgents, a multi-role, multi-LLM agent review framework, to enhance the review comment generation process. Additionally, we propose \textbf{ReviewBench}, a benchmark for evaluating the review comments generated by LLMs. Our experimental results on ReviewBench demonstrate that while existing LLMs exhibit a certain degree of potential for automating the review process, there remains a gap when compared to human-generated reviews. Moreover, our ReviewAgents framework further narrows this gap, outperforming advanced LLMs in generating review comments.

\end{abstract}
\section{Introduction}

Peer review of academic papers is a crucial component of the scholarly publishing system, as it ensures the quality of scientific research and facilitates the improvement of academic writing. However, with the rapid increase in the number of academic paper submissions, the traditional peer review process is confronted with challenges such as inefficiency and a limited number of reviewers, resulting in restricted feedback for authors and hindering the timely acquisition of review comments.

The reasoning capabilities of LLMs for complex tasks render them potentially valuable for providing feedback on academic papers \cite{liuReviewerGPTExploratoryStudy2023,zhaoWordsWorthNewborn2024,zhuangLargeLanguageModels2025}. This, to some extent, alleviates the pressure on reviewers and allows authors to independently use LLMs to review and revise their papers prior to formal submission. However, existing research and applications primarily focus on prompting LLMs to directly generate comments on the submitted papers \cite{zhouLLMReliableReviewer2024,darcyMARGMultiAgentReview2024}, which greatly simplifies the review process. This approach does not align with the cognitive process of human reviewers in real-world peer review, nor with the multi-step, multi-role nature of the peer review process. As a result, comments generated directly by LLMs are often difficult to align with the review comments provided by human reviewers. 
In fact, when human reviewers write their reviews, they are typically required to first summarize the main content of the paper, followed by enumerating its strengths and weaknesses, and ultimately reaching a review conclusion. After multiple reviewers have written their comments, the area chair (AC) synthesizes the feedback from all reviewers and provides a final summary as a meta-review. This process ensures that each reviewer fully understands the paper’s content, guaranteeing the accuracy, objectivity, and fairness of the review outcomes while minimizing the influence of individual reviewer biases on the final review comments.

To bridge the gap between LLM-based reviews and human review behavior, we have developed a multi-agent review framework, named ReviewAgents, which emulates the multi-step reasoning process of human reviewers. To train our agents, we collected a substantial amount of review comments from publicly available review platforms and structurally transcribed them, constructing the Review-CoT dataset, which guides the training of both reviewer agents and area chair agents. Specifically, Review-CoT contains 37,403 papers and 142,324 corresponding review comments and meta-reviews, collected from open peer review platforms. We utilized state-of-the-art large language models to transcribe the review comments into a stepwise reasoning format, aligning them with the cognitive process of human reviewers. Considering the timeliness of novelty assessments in review comments, we incorporated references to the most relevant papers, published up to the point of submission, into the dataset, and employed a relevant-paper-aware training method. This approach mirrors the process by which human reviewers evaluate novelty by retrieving relevant papers. The dataset is used to train both the reviewer and area chair agents. Upon completion of training, we employ ReviewAgents, a multi-agent framework, to simulate the review process by human reviewers and area chairs, utilizing multi-agent, multi-step review procedures to generate the final review comments. Unlike previous approaches, our ReviewAgents framework leverages the knowledge of relevant papers and fully emulating the paper evaluation process conducted by human experts.

To comprehensively assess the quality of generated review comments and their alignment with human reviewers' feedback, we propose ReviewBench, a benchmark specifically designed to evaluate the quality of automatically generated review comments. ReviewBench includes the latest papers and review comments from open platforms, ensuring that its data is not included in the pretraining datasets of most existing LLMs. We evaluate the capabilities of large language models in generating paper review comments across four dimensions.

The main contributions of our work are summarized as follows: 
\begin{enumerate}
    \item We introduce the Review-CoT dataset, which, to the best of our knowledge, is the first and largest dataset to simulate the human review thought process and includes relevant papers.
    \item We propose ReviewBench, a benchmark for the quantitative evaluation of review comments generated by LLMs. 
    \item By employing a relevant-paper-aware training method on the Review-CoT dataset, we construct ReviewAgents, a multi-step, multi-role framework for paper review that simulates the human review process. Experimental results demonstrate that our approach generates review comments closely aligned with those of human reviewers.
\end{enumerate}
\section{Related Work}

\subsection{Large Language Models for Paper Review}

Large language models (LLMs) have demonstrated significant potential in reviewing and comprehending complex texts \cite{liuReviewerGPTExploratoryStudy2023,zhaoWordsWorthNewborn2024,zhuangLargeLanguageModels2025,zhu2024llama}. Initial studies, which compared reviews generated by human reviewers and LLMs for academic papers, indicated that LLMs can produce comments with substantial overlap with those of human reviewers, contributing effectively to the peer review process \cite{robertsonGPT4SlightlyHelpful2023,liangCanLargeLanguage2023}. However, further research has shown that, despite the advanced capabilities of LLMs such as GPT-3.5 and GPT-4 in scoring input paper texts, the generated reviews still fall short of fully meeting human expectations \cite{zhouLLMReliableReviewer2024}. Researchers have fine-tuned LLMs to accommodate review requirements by constructing datasets from publicly available review comments \cite{kangDatasetPeerReviews2018,yuanCanWeAutomate2021,shenMReDMetaReviewDataset2022,dyckeNLPeerUnifiedResource2023,gaoReviewer2OptimizingReview2024,duLLMsAssistNLP2024}. Additionally, some studies have employed multi-turn dialogues \cite{tanPeerReviewMultiTurn2024} or prompted multiple LLMs to construct review systems \cite{darcyMARGMultiAgentReview2024}, providing comprehensive feedback on the entire text. 
AgentReview \cite{jinAgentReviewExploringPeer2024} attempts to simulate the interactions among different roles involved in the peer review process, generating review comments and discussion dialogues for each role. However, such interactions do not necessarily occur in real-world review scenarios, and its approach to generating review comments does not adhere to the cognitive patterns typically followed by human reviewers. In contrast to existing work, our approach aims to reconstruct the complete reasoning and evaluation process throughout the peer review workflow. We emphasize the in-depth reflection and multi-step, multi-role nature of real-world peer reviewing, thereby mitigating biases inherent in single-LLM-generated reviews and achieving better alignment with human reviewer expectations.

\subsection{Chain-of-thought Reasoning in Large Language Models}

Through chain-of-thought prompting \cite{weiChainofThoughtPromptingElicits2022}, LLMs are guided with a step-by-step reasoning process, enabling them to break down complex problems into a series of incremental reasoning steps \cite{chuNavigateEnigmaticLabyrinth2024a}. This decomposition enhances LLMs' performance in reasoning tasks, advancing structured prompting techniques \cite{qiaoPrismFrameworkDecoupling2024a,cesistaMultimodalStructuredGeneration2025}. Recent studies highlight that the systematization and structuring of reasoning significantly affect LLMs' performance, particularly their ability to perform multi-stage reasoning and identify key tasks at each stage. This can be achieved through independent language reasoning systems \cite{zhongEvaluationOpenAIO12024} or supervised fine-tuning \cite{xuLLaVACoTLetVision2025}. In this paper, we introduce structured reasoning to automated peer review, aligning LLM behavior with human review practices by dividing the review process into a three-stage structured procedure, improving review generation outcomes.

\section{Dataset and Benchmark}

\subsection{Structured Thinking in Review}

Given that paper review requires models to thoroughly read the manuscript, integrate external knowledge, and conduct an in-depth analysis of the paper’s strengths and weaknesses before assigning scores, a detailed stepwise reasoning process is essential. Inspired by the cognitive process of human reviewers, we decompose the review process of the intelligent review agent into three structured stages: \textbf{Summarization, Analysis, and Conclusion}, thereby aligning the automated review procedure with human-like reasoning. In the summarization stage, the agent is provided with the full text of the paper and tasked with concisely summarizing its main contributions and methodologies. During the analysis stage, the agent examines the strengths and weaknesses of the paper based on its content and relevant literature within the same domain, serving as the foundation for subsequent evaluation. In the conclusion stage, drawing upon the outcomes of the previous two stages, the agent formulates a final review verdict for the paper.

To facilitate this structured stepwise reasoning, we employ dedicated tags (e.g., <SUMMARY>...</SUMMARY>) to explicitly mark the beginning and end of each stage. These tags enable the model to maintain clarity throughout the entire reasoning process.

\subsection{Dataset Preparation}

We have defined a standardized data format for Review-CoT. Each data entry contains the following key information: a unique identifier, paper title, abstract, full-text content, submission conference, reviewer comments, Program Chair’s remarks, and the decision regarding paper acceptance. Additionally, the data records the titles and abstracts of the most relevant papers in the field up to the point of submission. With this information, the dataset samples we construct can be aligned as closely as possible with the information typically accessed by human reviewers during the review process, enabling the trained models to generate conclusions that are more consistent with those of human reviewers. 

\begin{figure}
    \centering
    \includegraphics[width=\linewidth]{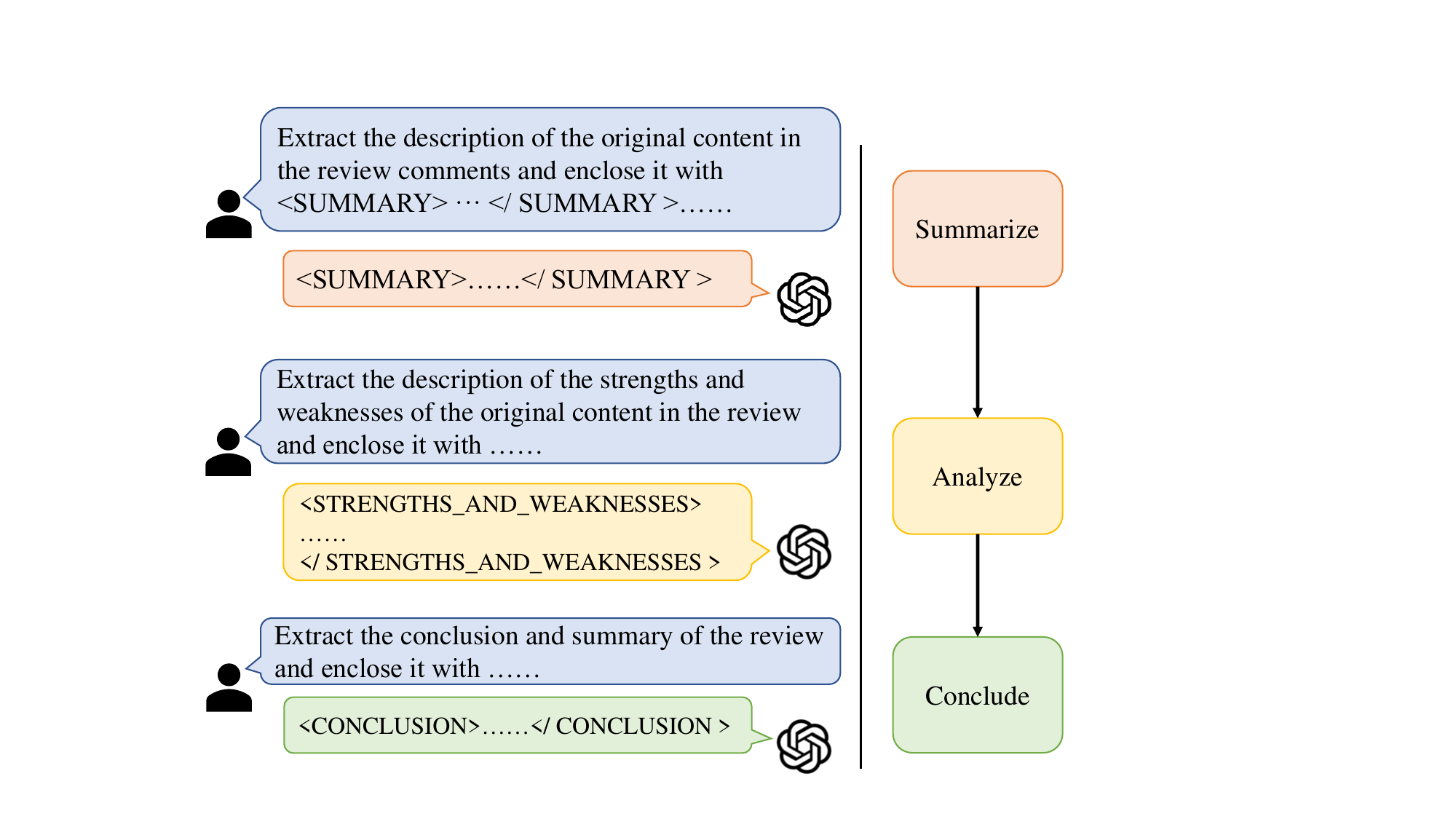}
    \caption{The process of transcribing the collected review comments using advanced LLMs according to a pre-defined structured thinking process}
    \label{fig:transcript}
\end{figure}

\subsubsection{Metadata Collection}

We crawled ICLR papers from 2017 to 2024 and NeurIPS papers from 2016 to 2024 from OpenReview\footnote{https://openreview.net/} and NeurIPS Proceedings\footnote{https://papers.nips.cc/}. For each paper, we adhere to the standardized data format we have defined, retaining as much metadata information as possible. Each paper includes the full text as well as the corresponding review comments. For the paper content, we use Scipdf\_Parser\footnote{https://github.com/titipata/scipdf\_parser} to parse the PDFs into structured JSON files.

\subsubsection{Transcription}

An examination of review comments within the collected metadata revealed that recent comments already follow a predefined structured review process, including a paper summary, analysis of strengths and weaknesses, and a final conclusion. While earlier comments contained these elements, they lacked explicit structural divisions. To address this, we employed advanced LLMs to transcribe the review comments according to the structured reasoning process (Figure \ref{fig:transcript}). This transcription reorganizes the paragraphs while preserving the original content and semantics, adding dedicated tags (e.g., <SUMMARY>...</SUMMARY>). This process transforms the review comments into a unified format suitable for model training.

\subsubsection{Relevant Paper Retrieval}

Human reviewers typically reference relevant literature to understand the field's current state to assess a paper's novelty, ensuring the timeliness of their review comments. The review comments in Review-CoT dataset, which evaluate content and novelty, reflect the reviewer's judgments during the review process. These judgments must be compared with the latest research to maintain timeliness, as outdated reviews cannot be generalized across time without considering recent findings. To enhance the reviewer's thought process and ensure timely comments, we use the Semantics Scholar API\footnote{https://www.semanticscholar.org/product/api} to retrieve papers in the same research domain as the reviewed paper, up to the submission date. We extract titles and abstracts from these papers and incorporate them into the dataset as references to assess the paper's novelty. Both the paper itself and the titles and abstracts of the retrieved relevant papers are encoded as vectors, and their similarity is computed. We retain the two papers with the highest similarity as the most relevant references for the reviewer during the review process.

\begin{figure*}
    \centering
    \includegraphics[width=0.96\linewidth]{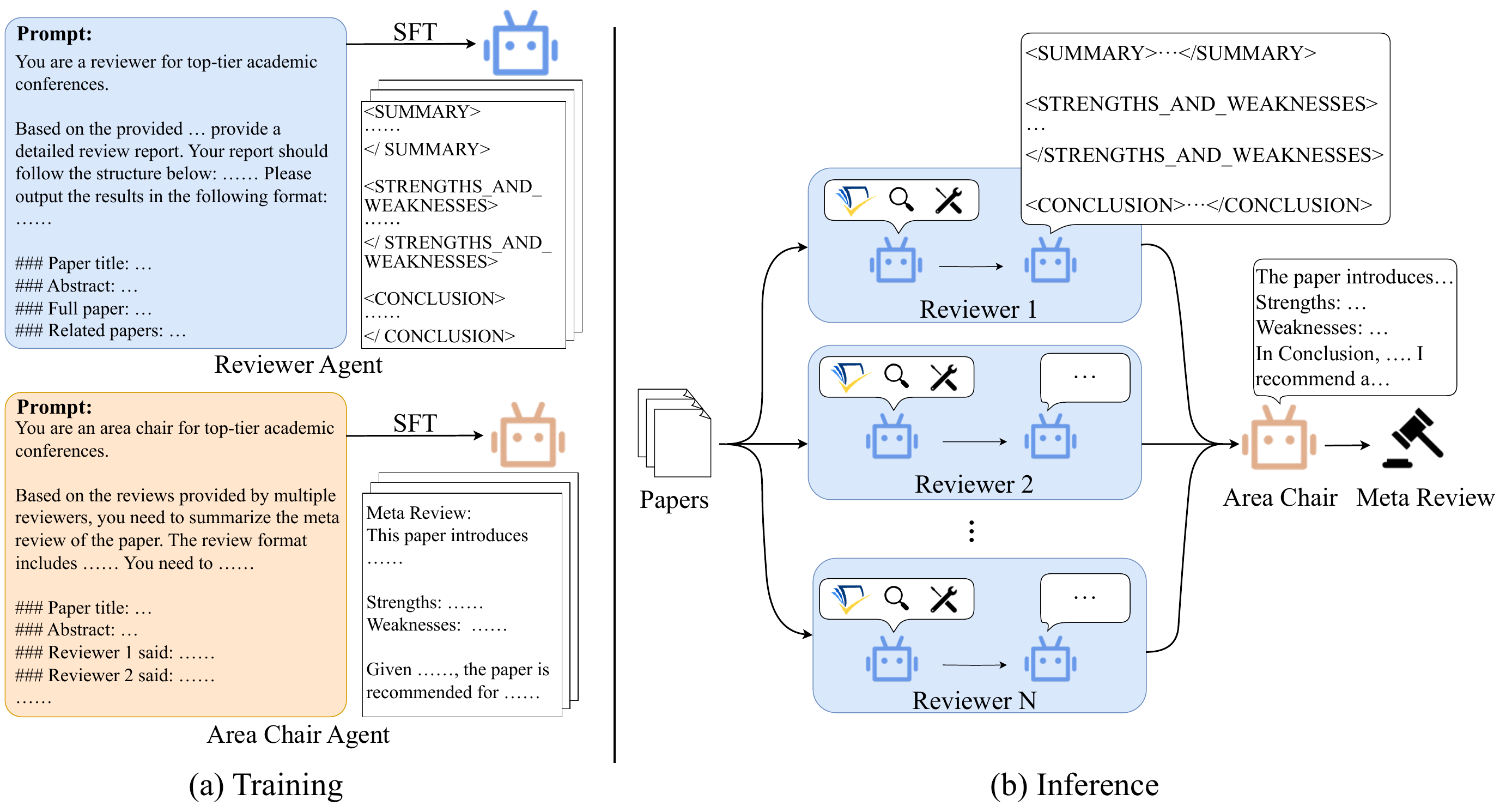}
    \caption{An overview of the ReviewAgents framework}
    \label{fig:framework}
\end{figure*}

\subsubsection{Dataset Details}

We have collected a total of 142,324 review comments for 37,403 papers from publicly available review platforms. Each review comment was transcribed into a structured format with appropriate delimiters added. For each paper, we included the titles and abstracts of the two most relevant papers, published up to the submission date, as references for the review process. 

For the training dataset of the Review Agent, the model's input consists of the original paper and the two relevant papers, with the output being the formatted review comments. For the Area Chair Agent's training dataset, the model's input is the output from the Review Agent, and the output is the paper's meta-review. 

Table \ref{table:dataset-comparison} presents a comparison between our dataset and existing review comment datasets, showing that our dataset outperforms the existing ones in both data volume and diversity.

\begin{table}
\centering
\resizebox{\columnwidth}{!}{%
\begin{tabular}{cccccc}
\hline
 &
  \# Papers &
  \# Reviews &
  \begin{tabular}[c]{@{}c@{}}Meta\\ Review\end{tabular} &
  \begin{tabular}[c]{@{}c@{}}Review\\ Process\end{tabular} &
  \begin{tabular}[c]{@{}c@{}}Related\\ Papers\end{tabular} \\ \hline
\begin{tabular}[c]{@{}c@{}}PeerRead\\ \cite{kangDatasetPeerReviews2018}\end{tabular}      & 3,006           & 10,770           & \xmark & \xmark & \xmark \\
\begin{tabular}[c]{@{}c@{}}ASAP-Review\\ \cite{yuanCanWeAutomate2021}\end{tabular}        & 8,877           & 28,119           & \xmark & \xmark & \xmark \\
\begin{tabular}[c]{@{}c@{}}MReD\\ \cite{shenMReDMetaReviewDataset2022}\end{tabular}       & 7,894           & 30,764           & \cmark & \xmark & \xmark \\
\begin{tabular}[c]{@{}c@{}}NLPeer\\ \cite{dyckeNLPeerUnifiedResource2023}\end{tabular}    & 5,672           & 11,515           & \xmark & \xmark & \xmark \\
\begin{tabular}[c]{@{}c@{}}ReviewCritique\\ \cite{duLLMsAssistNLP2024}\end{tabular}       & 120             & 440              & \xmark & \cmark & \xmark \\
\begin{tabular}[c]{@{}c@{}}Review2\\ \cite{gaoReviewer2OptimizingReview2024}\end{tabular} & 27,805          & 99,729           & \cmark & \xmark & \xmark \\ \hline
Review-CoT (ours)                                                                       & \textbf{37,403} & \textbf{142,324} & \cmark & \cmark & \cmark \\ \hline
\end{tabular}%
}
\caption{Comparison between Review-CoT and existing review comment datasets}
\label{table:dataset-comparison}
\end{table}

\subsection{ReviewBench}

Given that the evaluation of LLM-generated review comments is still limited to model scoring and human ratings, with a lack of standardized quantitative assessment criteria, we propose ReviewBench, a benchmark for evaluating LLM-generated review comments across multiple dimensions. Table \ref{table:benchmark} presents the evaluation dimensions and corresponding metrics included in ReviewBench. Our benchmark encompasses assessments of LLM-generated review comments in terms of language quality, semantic consistency, and sentiment consistency, as well as a Review Arena task to measure the model's ability to generate review comments. Further details on the evaluation metrics can be found in 
Appendix \ref{appendix:metrics}.

\subsubsection{Language Diversity}

The language diversity dimension is evaluated using Distinct \cite{liDiversityPromotingObjectiveFunction2016} and Inverse Self-BLEU as metrics. \textbf{Distinct} measures the proportion of non-redundant n-grams in the generated text relative to the total number of n-grams, reflecting the degree of repetition in the text and serving as an indicator of its diversity. Based on Self-BLEU \cite{zhuTexygenBenchmarkingPlatform2018}, \textbf{Inverse Self-BLEU} is inversely correlated with the BLEU score between n-grams in the text. The higher the Inverse Self-BLEU score, the more diverse the text generation, facilitating the construction of a composite score in conjunction with other metrics. The calculation formula is given by: $\text{Inverse Self-BLEU} = 2 - \text{Self-BLEU}$.

\subsubsection{Semantic Consistency}

Semantic consistency is employed to assess the degree of semantic alignment between the review comments generated by the model and those written by human reviewers, using the ROUGE-1, ROUGE-L, and SPICE metrics. ROUGE \cite{linROUGEPackageAutomatic2004} is a widely used metric for evaluating the overlap between generated text and reference text. \textbf{ROUGE-1} measures the overlap of individual words between the generated and reference texts, while \textbf{ROUGE-L} evaluates the longest common subsequence (LCS) between them. The \textbf{SPICE} metric \cite{andersonSPICESemanticPropositional2016}, originally designed to evaluate the consistency of image captions, measures semantic consistency by comparing part-of-speech and semantic structures, assessing whether the generated text aligns semantically with the reference text.

\subsubsection{Sentiment Consistency}

Sentiment consistency is employed to measure the overall alignment of sentiment and attitudinal orientation between the review comments generated by the model and those written by human reviewers, specifically assessing whether both are predominantly positive or negative \cite{wangSentimentAnalysisPeer2018}. The sentiment tendencies of both model-generated and human review comments are extracted using sentiment analysis models based on BERT \cite{devlinBERTPretrainingDeep2019} and VADER \cite{huttoVADERParsimoniousRuleBased2014}, respectively. The distance between these tendencies is then evaluated as the sentiment consistency scores, \textbf{BERTScore} and \textbf{VADER Score}.

\subsubsection{Review Arena}

The open nature of peer review presents significant challenges for automated evaluation. Previous studies have shown that direct scoring methods using LLMs do not align well with human preferences \cite{liangCanLargeLanguage2023}. In contrast, LLMs perform better in ranking tasks \cite{liChainIdeasRevolutionizing2024}. To obtain reliable evaluations, we propose Review Arena, which employs a tournament-style pairwise evaluation system to compare pairs of review comments. To mitigate the influence of hallucination, we use human review comments as a reference and require LLM judges to rank a pair of model-generated review comments based on their similarity in content and form to the human review comments. Each pair of review comments undergoes two evaluations through reverse sorting to minimize positional bias.

\begin{table}
\centering
\small
\begin{tabular}{cc}
\hline
Dimensions                             & Metrics               \\ \hline
\multirow{2}{*}{Language Diversity}     & $\text{Distinct}_4$                \\
                                      & Inverse Self-BLEU@4                    \\ \hline
\multirow{3}{*}{Semantic Consistency}  & ROUGE-1                        \\
                                      & ROUGE-L                        \\
                                      & SPICE                        \\ \hline
\multirow{2}{*}{Sentiment Consistency} & BERTScore                    \\
                                      & VADER Score                  \\ \hline
Review Arena                       &  Win Rate           \\ \hline
\end{tabular}%
\caption{Evaluation dimensions and corresponding metrics in ReviewBench}
\label{table:benchmark}
\end{table}

\subsubsection{Statistical Information}

We selected 100 papers from ICLR 2024 and NeurIPS 2024 as the test set for our benchmark. ICLR 2024 and NeurIPS 2024 were published in February and September of 2024, respectively, both of which postdate the knowledge cutoff times of most LLMs' pretraining datasets\footnote{The knowledge cutoff time for the GPT-4o and Llama-3.1 is October 2023 and December 2023, respectively.}, ensuring that the test data was not used during the pretraining phase. These papers were removed from the Review-CoT training dataset. To better simulate the real review process, the selected benchmark follows an acceptance-to-rejection ratio of 3:7, which closely approximates the acceptance rate of these two conferences.

\begin{table*}
\centering
\small
\resizebox{0.99\textwidth}{!}{%
\begin{tabular}{c|ccccccc|c}
\hline
\multirow{2}{*}{Model} &
  \multicolumn{2}{c}{Language Diversity $\uparrow$} &
  \multicolumn{3}{c}{Semantic Consistency $\uparrow$} &
  \multicolumn{2}{c|}{\begin{tabular}[c]{@{}c@{}}Sentiment\\ Consistency\end{tabular} $\uparrow$} &
  \multirow{2}{*}{Overall $\uparrow$}\\ \cline{2-8}
 &
   $\text{Distinct}_4$ &
  \begin{tabular}[c]{@{}c@{}}  Inverse\\ Self-BLEU@4\end{tabular} &
   ROUGE-1 &
   ROUGE-L &
   SPICE &
  \begin{tabular}[c]{@{}c@{}} BERT\\ Score\end{tabular} &
  \begin{tabular}[c]{@{}c@{}} VADER\\ Score\end{tabular} &
   \\ \hline
Human &
  98.88 &
  89.84 &
  100.00 &
  100.00 &
  100.00 &
  100.00 &
  100.00 &
  98.39 \\ \hline
\multicolumn{9}{c}{\textit{API-based LLMs}} \\ \hline
Claude-3.5-sonnet &
  98.50 &
  \textbf{84.39} &
  39.93 &
  16.26 &
  14.78 &
  33.65 &
  {\ul 86.29} &
  {\ul 53.40} \\
Deepseek-R1 &
  \textbf{99.37} &
  77.64 &
  37.22 &
  14.40 &
  13.57 &
  \textbf{48.13} &
  83.09 &
  53.35 \\
GPT-4o &
  {\ul 99.26} &
  {\ul 82.48} &
  {\ul 41.35} &
  16.62 &
  14.64 &
  30.55 &
  84.32 &
  52.75 \\
Deepseek-V3 &
  95.51 &
  69.36 &
  38.06 &
  16.57 &
  14.58 &
  30.20 &
  85.70 &
  50.00 \\
GLM-4 &
  89.80 &
  64.33 &
  40.69 &
  {\ul 18.29} &
  {\ul 15.33} &
  30.98 &
  \textbf{86.79} &
  49.46 \\ \hline
\multicolumn{9}{c}{\textit{Open-source LLMs}} \\ \hline
Llama-3.1-70B-Instruct &
  87.58 &
  60.94 &
  40.53 &
  17.83 &
  15.24 &
  35.56 &
  84.02 &
  48.81 \\
Qwen-2.5-72B-Chat &
  94.72 &
  64.93 &
  29.97 &
  15.56 &
  15.18 &
  26.42 &
  84.66 &
  47.35 \\
Llama-3.1-8B-Instruct &
  82.98 &
  55.70 &
  30.39 &
  13.90 &
  12.95 &
  26.23 &
  81.00 &
  43.31 \\ \hline
Ours &
  96.57 &
  77.60 &
  \textbf{42.88} &
  \textbf{19.27} &
  \textbf{15.75} &
  {\ul 44.71} &
  86.26 &
  \textbf{54.72} \\ \hline
\end{tabular}%
}
\caption{Comparison of the diversity and consistency metric scores}
\label{table:metrics}

\end{table*}

\section{Methodology}

\subsection{Framework Overview}

Figure \ref{fig:framework} illustrates an overview of the ReviewAgents framework we propose. During the training phase, we employ a training strategy that incorporates awareness of relevant papers and a structured reasoning approach to train the reviewer agent, while using meta-review to train the area chair agent, enabling the model to simulate the human peer review process. In the inference phase, the reviewer, trained with awareness of relevant papers, generates review comments by considering relevant papers retrieved from the literature database. The area chair then synthesizes multiple review comments to form the final review conclusion.

\subsection{Relevant-paper-aware Training}

The timeliness of review comments is reflected in the relevant papers that have been published up until the time of the review. In the Review-CoT training dataset, relevant papers are provided as references for the review process. During training, the model learns how to integrate these relevant papers into the generation of review comments, thereby incorporating timeliness information and closely replicating the reasoning process employed by human reviewers when crafting their feedback. To ensure that the model accesses the relevant paper information without exceeding the context window, we input the titles and abstracts of the relevant papers as prompts during training.

\subsection{Multi-step Multi-role Review Generation}

After training LLM agents for different roles, we construct a review framework that emulates the human review process. The paper is initially assigned to multiple distinct reviewers, each of whom utilizes a retrieval API to search for relevant papers in the literature database as references for their review. Each reviewer then follows a structured reasoning process in sequence—SUMMARY, ANALYZE, and CONCLUDE—to generate their review comments. Subsequently, the review comments are submitted to the area chair agent for aggregation and synthesis, eliminating any biases present in the individual reviewer's feedback. The final review comments, provided by the area chair, serve as the ultimate output of the review process.
\begin{figure}
    \centering
    \includegraphics[width=\linewidth]{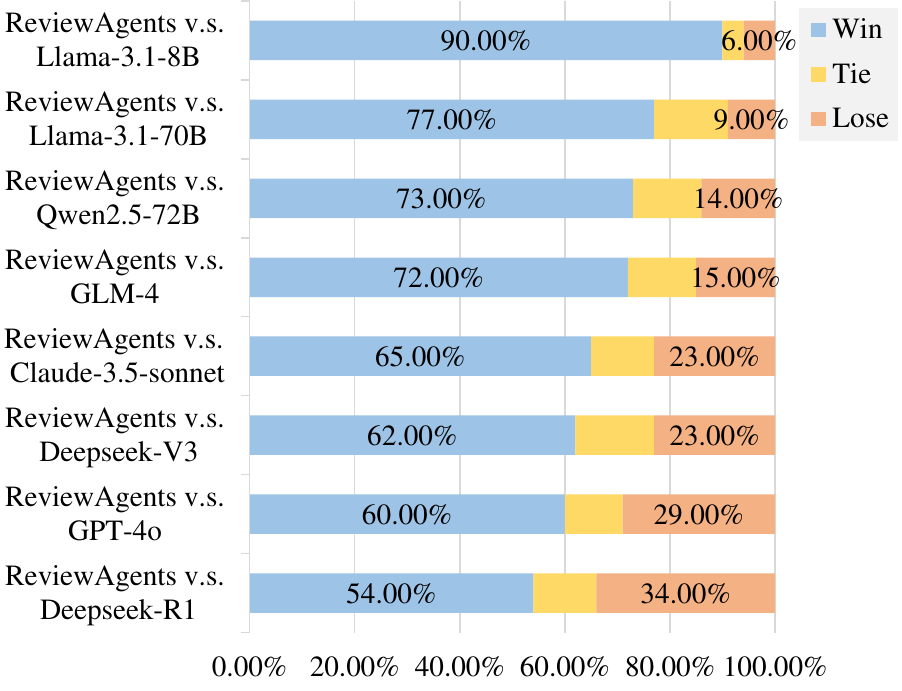}
    \caption{Win rates of ReviewAgents in Review Arena}
    \label{fig:winrate}
\end{figure}

\section{Experiments}
\subsection{Experimental Setup}

We evaluated our method on ReviewBench, selecting advanced open-source models
and closed-source models
for comparison, evaluating the meta-reviews generated by different models. The evaluation metrics, Distinct and Inverse Self-BLEU, are computed using 4-grams, denoted as $\text{Distinct}_4$ and Inverse Self-BLEU@4, respectively. For ROUGE-1 and ROUGE-L, their F1-scores are computed. For the diversity and consistency metrics, including Language Diversity, Semantic Consistency, and Sentiment Consistency, we compute their overall scores. Detailed implementation specifics and prompt templates are provided in 
Appendix \ref{appendix:implement} and \ref{appendix:prompt}.

\subsection{Main Results}
\subsubsection{Diversity and Consistency}

Table \ref{table:metrics} presents a comparison of the diversity and consistency metric scores of our ReviewAgents with those of other models on ReviewBench. It is evident that, despite being constrained by the model's parameter size, the proposed ReviewAgents method exhibits slight limitations in language diversity. However, the review comments generated by ReviewAgents outperform larger parameter open-source models in both semantic consistency and viewpoint alignment, achieving results comparable to or exceeding those of advanced closed-source models. In terms of overall metrics, our approach achieves state-of-the-art results, validating the comprehensiveness and effectiveness of our method.

\begin{table*}
\centering
\small
\resizebox{\textwidth}{!}{%
\begin{tabular}{c|ccccccc|c}
\hline
\multirow{2}{*}{\begin{tabular}[c]{@{}c@{}}$N$ of \\ Reviewers\end{tabular}} &
  \multicolumn{2}{c}{Language Diversity $\uparrow$} &
  \multicolumn{3}{c}{Semantic Consistency $\uparrow$} &
  \multicolumn{2}{c|}{\begin{tabular}[c]{@{}c@{}}Sentiment\\ Consistency\end{tabular} $\uparrow$} &
  \multirow{2}{*}{Overall} \\ \cline{2-8}
 &
  Distinct4 &
  \begin{tabular}[c]{@{}c@{}}Inverse\\ Self-BLEU@4\end{tabular} &
  ROUGE-1 &
  ROUGE-L &
  SPICE &
  \begin{tabular}[c]{@{}c@{}}BERT\\ Score\end{tabular} &
  \begin{tabular}[c]{@{}c@{}}VADER\\ Score\end{tabular} &
   \\ \hline
1 & \textbf{97.78} & \textbf{82.29} & \textbf{43.20} & \textbf{19.41} & 15.66          & 31.67          & 84.83          & 53.55          \\
2 & {\ul 96.69}    & {\ul 78.08}    & 42.53          & {\ul 19.37}    & \textbf{15.82} & 44.49          & 84.31          & 54.47          \\
3 & 96.57          & 77.60          & 42.88          & 19.27          & {\ul 15.75}    & \textbf{44.71} & \textbf{86.26} & \textbf{54.72} \\
4 & 96.49          & 76.33          & {\ul 43.13}    & 19.27          & 15.71          & {\ul 44.52}    & {\ul 86.03}    & {\ul 54.50}    \\
5 & 96.16          & 75.95          & 42.23          & 19.04          & 15.55          & 42.98          & 85.85          & 53.97          \\
6 & 95.81          & 75.24          & 42.25          & 19.08          & 15.67          & 43.39          & 85.91          & 53.91          \\ \hline
\end{tabular}%
}
\caption{Ablation study on the number of reviewer agents}
\label{table:ablation}
\end{table*}

\subsubsection{Review Arena Results}

Figure \ref{fig:winrate} illustrates the win rates of ReviewAgents compared to other models in the Review Arena. ReviewAgents achieved a higher win rate than closed-source LLMs and significantly outperformed open-source LLMs, further validating the effectiveness of our approach in generating review comments that align closely with human feedback.

\begin{table}[!t]
\centering
\small
\resizebox{\columnwidth}{!}{%
\begin{tabular}{c|ccccc|c}
\hline
                       & SV            & CO            & C             & CHR           & PC            & Average       \\ \hline
ReviewAgents           & 3.46          & \textbf{4.40} & 3.43          & \textbf{3.99} & {\ul 3.85}    & \textbf{3.83} \\
Deepseek-R1            & 3.65          & {\ul 4.06}    & {\ul 3.94}    & {\ul 3.94}    & 3.45          & {\ul 3.81}    \\
Deepseek-V3            & 3.61          & 3.85          & 3.62          & 3.82          & \textbf{3.92} & 3.76          \\
GPT-4o                 & \textbf{3.80} & 3.96          & \textbf{4.00} & 3.84          & 2.83          & 3.69          \\
Claude-3.5-sonnet      & 3.58          & 3.82          & 3.78          & 3.09          & 3.72          & 3.60          \\
Qwen-2.5-72B-Chat      & 3.48          & 3.79          & 3.46          & 2.76          & 3.11          & 3.32          \\
GLM-4                  & 2.80          & 3.57          & 3.77          & 3.07          & 2.75          & 3.19          \\
Llama-3.1-70B-Instruct & {\ul 3.66}    & 3.50          & 3.23          & 2.53          & 3.00          & 3.18          \\
Llama-3.1-8B-Instruct  & 2.68          & 3.32          & 2.85          & 2.02          & 2.57          & 2.69          \\ \hline
\end{tabular}%
}
\caption{The results of the human evaluation}
\label{tab:human-eval-results}
\end{table}

\subsubsection{Human Evaluation Results}

We engaged five AI researchers with submission and reviewing experience at top-tier AI conferences such as ICLR, NeurIPS, and *CL to serve as evaluators. They assessed and scored the review comments generated by different LLM-based methods. The evaluation dimensions included \textit{Soundness \& Validity (SV)}, \textit{Clarity \& Organization (CO)}, \textit{Constructiveness (C)}, \textit{Consistency with Human Review (CHR)}, and \textit{Position Clarity (PC)}. Each dimension was rated on a scale from 1 to 5. Detailed descriptions of each evaluation dimension can be found in Appendix \ref{questionnaire}.

Table \ref{tab:human-eval-results} presents the results of the human evaluation. ReviewAgents achieved higher average scores compared to proprietary LLMs, with particularly notable improvements in the dimensions of \textit{Clarity \& Organization (CO)} and \textit{Consistency with Human Review (CHR)}. These findings further substantiate that the reviews generated by our approach align more closely with the expectations of human reviewers.

\subsection{Ablation Study on Number of Reviewer Agents}

The number of agents in multi-agent systems directly affects the final outcome \cite{zhangExploringCollaborationMechanisms2024}. To determine the optimal number of reviewers within the ReviewAgents framework, we examined the performance of review comments generated with different reviewer counts. Table \ref{table:ablation} shows the impact of varying values of $N$ on the generated review comments. The results indicate that when $N=3$, the generated review comments align most closely with human reviewers' feedback. Further experiments reveal that fewer reviewers yield more diverse comments, while an excessive number of reviewers does not improve alignment with human opinions. This may be due to the fact that, in the training dataset, the average number of review comments per paper is between 3 and 4. As a result, too many review comments during inference introduce opinion discrepancies and extended context, negatively affecting the area chair agent's judgment.

\subsection{Discussion and Case Study}

As shown in Table \ref{table:metrics}, the improvements of ReviewAgents over other methods primarily stem from enhanced semantic and sentiment similarity with human review comments. This improvement is primarily due to the structured generation process of review comments, which mirrors the stages of human peer review. As a result, each LLM stage generates review comments through a reasoning process closely aligned with human reviewers. Additionally, the final meta-review is derived from aggregating comments from multiple reviewer agents, reducing individual reviewer biases.

\begin{figure}[!t]
    \centering
    \includegraphics[width=\linewidth]{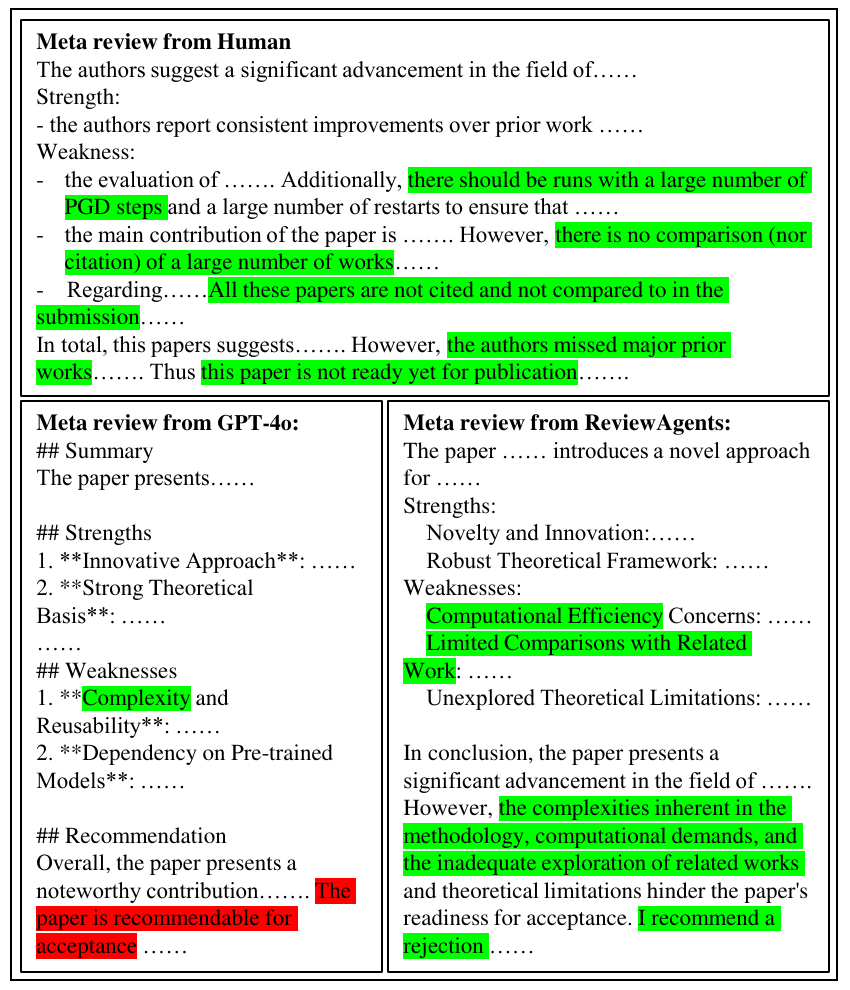}
    \caption{A Case Study. The \textcolor{green}{green} highlighted sections represent matching opinions, while the \textcolor{red}{red} highlighted sections represent opposing decisions.}
    \label{fig:case-study}
\end{figure}

Figure \ref{fig:case-study} presents a case study. Compared to comments from advanced LLMs, ReviewAgents' comments show greater overlap with human reviewers' comments. For example, both address issues such as insufficient comparison and citation, highlighting the effectiveness of our relevant-paper-aware training approach. While GPT-4o tends to generate more positive comments, ReviewAgents, through stepwise analysis of strengths and weaknesses, produces rejection comments aligned with human reviewers' assessments. We provide a complete case of the peer review comment generation process in the 
Appendix \ref{appendix:case}.

\section{Conclusion}

In this paper, we aim to enhance the ability of LLMs to generate review comments, thereby advancing the automation of academic paper review. To achieve this objective, we collected papers and review comments and constructed the Review-CoT dataset, which includes the original papers, review comments with structured thinking processes, area chair comments, and relevant papers. Building on this, we employed a relevant-paper-aware method for fine-tuning LLMs and introduced ReviewAgents, a multi-agent review system that simulates the human review process, incorporating relevant paper retrieval, structured thinking, and a multi-role, multi-step approach. To comprehensively evaluate the capacity of LLMs to generate review comments, we developed ReviewBench, a benchmark comprising four testing dimensions. Experimental results demonstrate that our ReviewAgents surpass advanced LLMs, such as GPT-4o and Deepseek-R1.

\section{Limitations}
In this section, we discuss some of the limitations of our work.

\subsection{Limited Research Area}

Thanks to the openness of paper reviews in the AI field, our current work constructs datasets using review comments from AI conference papers. As a result, our approach to generating review comments is currently limited to the AI domain. In the future, we plan to collect papers and review comments from additional fields, thereby constructing datasets that encompass a broader range of disciplines, which will enhance the generalizability of our method across diverse research areas.

\subsection{Disjoint Process for Review Generation}

Our approach generates the final review comments through a multi-step generation process. During training, the reviewer agent and area chair agent are trained separately, which results in a disconnection between the training process and the inference phase. This may cause the input distribution for the area chair agent during the inference phase to differ from that during training. A potential improvement could involve performing joint training after separate supervised fine-tuning in both stages.

\subsection{Limited Context Length}

Due to the limitations imposed by the context window of large language models, ReviewAgents is constrained by a maximum paper length. Papers that exceed this length may experience truncation, leading to potential biases in the final review comments. Expanding the context window could result in increased training and inference costs. A potential solution to this issue is the batching of papers for processing.

\section*{Ethical Statement}

This study constructs our dataset and benchmark using publicly available research paper data. All data used in this research is sourced from openly accessible academic repositories and databases, and no personal or sensitive information of authors or research participants is involved. Data processing and analysis adhered to the relevant terms of use of these data sources, and no privacy or confidentiality issues arose during the data collection process. The authors declare no conflicts of interest, and all analyses were conducted using automated methods, with complete transparency in the algorithms and data processing procedures. This research follows the ethical guidelines for the responsible use of data in academic research. 

The potential risk of our research is the abuse of automatic peer review, which could lead to a decline in the quality of review comments and irresponsible feedback. We emphasize that our work is not intended to replace human reviewers. Instead, our approach aims to enhance the efficiency of the review process while maintaining scientific rigor and alignment with human evaluation standards, rather than replacing human reviewers. The proposed method is designed to serve as an auxiliary tool—integrating LLM-generated reviews as a component of the overall review pipeline—to offer complementary perspectives to traditional expert assessments. In essence, we are exploring how LLMs can augment, rather than supplant, the irreplaceable expertise and discernment of human reviewers. Additionally, our work can serve as a valuable tool for authors to self-edit their papers and as an aid for human reviewers. The results produced by the model should not be misconstrued as definitive and genuine comments on the respective papers.

All annotators were carefully recruited and reasonably compensated for their contributions. They received substantial remuneration to ensure that their time and expertise were fairly valued. Annotators were fully informed of the research objectives and the intended use of the data. All annotators operated under strict institutional guidelines to ensure that data handling adhered to ethical standards, and confidentiality was maintained throughout the annotation and evaluation processes.

\bibliography{acl_latex}

\begin{thebibliography}{36}
\providecommand{\natexlab}[1]{#1}

\bibitem[{Anderson et~al.(2016)Anderson, Fernando, Johnson, and Gould}]{andersonSPICESemanticPropositional2016}
Peter Anderson, Basura Fernando, Mark Johnson, and Stephen Gould. 2016.
\newblock \href {https://doi.org/10.48550/arXiv.1607.08822} {{{SPICE}}: {{Semantic Propositional Image Caption Evaluation}}}.
\newblock \emph{Preprint}, arXiv:1607.08822.

\bibitem[{Cesista(2025)}]{cesistaMultimodalStructuredGeneration2025}
Franz~Louis Cesista. 2025.
\newblock \href {https://doi.org/10.48550/arXiv.2406.11403} {Multimodal {{Structured Generation}}: {{CVPR}}'s 2nd {{MMFM Challenge Technical Report}}}.
\newblock \emph{Preprint}, arXiv:2406.11403.

\bibitem[{Chu et~al.(2024)Chu, Chen, Chen, Yu, He, Wang, Peng, Liu, Qin, and Liu}]{chuNavigateEnigmaticLabyrinth2024a}
Zheng Chu, Jingchang Chen, Qianglong Chen, Weijiang Yu, Tao He, Haotian Wang, Weihua Peng, Ming Liu, Bing Qin, and Ting Liu. 2024.
\newblock \href {https://doi.org/10.18653/v1/2024.acl-long.65} {Navigate through {{Enigmatic Labyrinth A Survey}} of {{Chain}} of {{Thought Reasoning}}: {{Advances}}, {{Frontiers}} and {{Future}}}.
\newblock In \emph{Proceedings of the 62nd {{Annual Meeting}} of the {{Association}} for {{Computational Linguistics}} ({{Volume}} 1: {{Long Papers}})}, pages 1173--1203, Bangkok, Thailand. Association for Computational Linguistics.

\bibitem[{D'Arcy et~al.(2024)D'Arcy, Hope, Birnbaum, and Downey}]{darcyMARGMultiAgentReview2024}
Mike D'Arcy, Tom Hope, Larry Birnbaum, and Doug Downey. 2024.
\newblock \href {https://doi.org/10.48550/arXiv.2401.04259} {{{MARG}}: {{Multi-Agent Review Generation}} for {{Scientific Papers}}}.
\newblock \emph{Preprint}, arXiv:2401.04259.

\bibitem[{Devlin et~al.(2019)Devlin, Chang, Lee, and Toutanova}]{devlinBERTPretrainingDeep2019}
Jacob Devlin, Ming-Wei Chang, Kenton Lee, and Kristina Toutanova. 2019.
\newblock \href {https://doi.org/10.18653/v1/N19-1423} {{{BERT}}: {{Pre-training}} of {{Deep Bidirectional Transformers}} for {{Language Understanding}}}.
\newblock In \emph{Proceedings of the 2019 {{Conference}} of the {{North American Chapter}} of the {{Association}} for {{Computational Linguistics}}: {{Human Language Technologies}}, {{Volume}} 1 ({{Long}} and {{Short Papers}})}, pages 4171--4186, Minneapolis, Minnesota. Association for Computational Linguistics.

\bibitem[{Du et~al.(2024)Du, Wang, Zhao, Deng, Liu, Lou, Zou, Venkit, Zhang, Srinath, Zhang, Gupta, Li, Li, Wang, Liu, Liu, Gao, Xia, Xing, Cheng, Wang, Su, Shah, Guo, Gu, Li, Wei, Wang, Cheng, Ranathunga, Fang, Fu, Liu, Huang, Blanco, Cao, Zhang, Yu, and Yin}]{duLLMsAssistNLP2024}
Jiangshu Du, Yibo Wang, Wenting Zhao, Zhongfen Deng, Shuaiqi Liu, Renze Lou, Henry~Peng Zou, Pranav~Narayanan Venkit, Nan Zhang, Mukund Srinath, Haoran~Ranran Zhang, Vipul Gupta, Yinghui Li, Tao Li, Fei Wang, Qin Liu, Tianlin Liu, Pengzhi Gao, Congying Xia, Chen Xing, Jiayang Cheng, Zhaowei Wang, Ying Su, Raj~Sanjay Shah, Ruohao Guo, Jing Gu, Haoran Li, Kangda Wei, Zihao Wang, Lu~Cheng, Surangika Ranathunga, Meng Fang, Jie Fu, Fei Liu, Ruihong Huang, Eduardo Blanco, Yixin Cao, Rui Zhang, Philip~S. Yu, and Wenpeng Yin. 2024.
\newblock \href {https://doi.org/10.48550/arXiv.2406.16253} {{{LLMs Assist NLP Researchers}}: {{Critique Paper}} ({{Meta-}}){{Reviewing}}}.
\newblock \emph{Preprint}, arXiv:2406.16253.

\bibitem[{Dubey et~al.(2024)Dubey, Jauhri, Pandey, Kadian, Al-Dahle, Letman, Mathur, Schelten, Yang, Fan et~al.}]{dubey2024llama}
Abhimanyu Dubey, Abhinav Jauhri, Abhinav Pandey, Abhishek Kadian, Ahmad Al-Dahle, Aiesha Letman, Akhil Mathur, Alan Schelten, Amy Yang, Angela Fan, et~al. 2024.
\newblock \href {https://doi.org/10.48550/arXiv.2407.21783} {The llama 3 herd of models}.
\newblock \emph{Preprint}, arXiv:2407.21783.

\bibitem[{Dycke et~al.(2023)Dycke, Kuznetsov, and Gurevych}]{dyckeNLPeerUnifiedResource2023}
Nils Dycke, Ilia Kuznetsov, and Iryna Gurevych. 2023.
\newblock \href {https://doi.org/10.18653/v1/2023.acl-long.277} {{{NLPeer}}: {{A Unified Resource}} for the {{Computational Study}} of {{Peer Review}}}.
\newblock In \emph{Proceedings of the 61st {{Annual Meeting}} of the {{Association}} for {{Computational Linguistics}} ({{Volume}} 1: {{Long Papers}})}, pages 5049--5073, Toronto, Canada. Association for Computational Linguistics.

\bibitem[{Gao et~al.(2024)Gao, Brantley, and Joachims}]{gaoReviewer2OptimizingReview2024}
Zhaolin Gao, Kiant{\'e} Brantley, and Thorsten Joachims. 2024.
\newblock \href {https://doi.org/10.48550/arXiv.2402.10886} {Reviewer2: {{Optimizing Review Generation Through Prompt Generation}}}.
\newblock \emph{Preprint}, arXiv:2402.10886.

\bibitem[{Guo et~al.(2025)Guo, Yang, Zhang, Song, Zhang, Xu, Zhu, Ma, Wang, Bi et~al.}]{guo2025deepseek}
Daya Guo, Dejian Yang, Haowei Zhang, Junxiao Song, Ruoyu Zhang, Runxin Xu, Qihao Zhu, Shirong Ma, Peiyi Wang, Xiao Bi, et~al. 2025.
\newblock Deepseek-r1: Incentivizing reasoning capability in llms via reinforcement learning.
\newblock \emph{arXiv preprint arXiv:2501.12948}.

\bibitem[{Hutto and Gilbert(2014)}]{huttoVADERParsimoniousRuleBased2014}
C.~Hutto and Eric Gilbert. 2014.
\newblock \href {https://doi.org/10.1609/icwsm.v8i1.14550} {{{VADER}}: {{A Parsimonious Rule-Based Model}} for {{Sentiment Analysis}} of {{Social Media Text}}}.
\newblock \emph{Proceedings of the International AAAI Conference on Web and Social Media}, 8(1):216--225.

\bibitem[{Jin et~al.(2024)Jin, Zhao, Wang, Chen, Zhu, Xiao, and Wang}]{jinAgentReviewExploringPeer2024}
Yiqiao Jin, Qinlin Zhao, Yiyang Wang, Hao Chen, Kaijie Zhu, Yijia Xiao, and Jindong Wang. 2024.
\newblock \href {https://doi.org/10.48550/arXiv.2406.12708} {{{AgentReview}}: {{Exploring Peer Review Dynamics}} with {{LLM Agents}}}.
\newblock \emph{Preprint}, arXiv:2406.12708.

\bibitem[{Kang et~al.(2018)Kang, Ammar, Dalvi, {van Zuylen}, Kohlmeier, Hovy, and Schwartz}]{kangDatasetPeerReviews2018}
Dongyeop Kang, Waleed Ammar, Bhavana Dalvi, Madeleine {van Zuylen}, Sebastian Kohlmeier, Eduard Hovy, and Roy Schwartz. 2018.
\newblock \href {https://doi.org/10.18653/v1/N18-1149} {A {{Dataset}} of {{Peer Reviews}} ({{PeerRead}}): {{Collection}}, {{Insights}} and {{NLP Applications}}}.
\newblock In \emph{Proceedings of the 2018 {{Conference}} of the {{North American Chapter}} of the {{Association}} for {{Computational Linguistics}}: {{Human Language Technologies}}, {{Volume}} 1 ({{Long Papers}})}, pages 1647--1661, New Orleans, Louisiana. Association for Computational Linguistics.

\bibitem[{Li et~al.(2016)Li, Galley, Brockett, Gao, and Dolan}]{liDiversityPromotingObjectiveFunction2016}
Jiwei Li, Michel Galley, Chris Brockett, Jianfeng Gao, and Bill Dolan. 2016.
\newblock \href {https://doi.org/10.18653/v1/N16-1014} {A {{Diversity-Promoting Objective Function}} for {{Neural Conversation Models}}}.
\newblock In \emph{Proceedings of the 2016 {{Conference}} of the {{North American Chapter}} of the {{Association}} for {{Computational Linguistics}}: {{Human Language Technologies}}}, pages 110--119, San Diego, California. Association for Computational Linguistics.

\bibitem[{Li et~al.(2024)Li, Xu, Guo, Zhao, Li, Yuan, Zhang, Jiang, Xin, Dang, Zhao, Rong, Feng, and Bing}]{liChainIdeasRevolutionizing2024}
Long Li, Weiwen Xu, Jiayan Guo, Ruochen Zhao, Xingxuan Li, Yuqian Yuan, Boqiang Zhang, Yuming Jiang, Yifei Xin, Ronghao Dang, Deli Zhao, Yu~Rong, Tian Feng, and Lidong Bing. 2024.
\newblock \href {https://arxiv.org/abs/2410.13185} {Chain of {{Ideas}}: {{Revolutionizing Research Via Novel Idea Development}} with {{LLM Agents}}}.
\newblock \emph{Preprint}, arXiv:2410.13185.

\bibitem[{Liang et~al.(2023)Liang, Zhang, Cao, Wang, Ding, Yang, Vodrahalli, He, Smith, Yin, McFarland, and Zou}]{liangCanLargeLanguage2023}
Weixin Liang, Yuhui Zhang, Hancheng Cao, Binglu Wang, Daisy Ding, Xinyu Yang, Kailas Vodrahalli, Siyu He, Daniel Smith, Yian Yin, Daniel McFarland, and James Zou. 2023.
\newblock \href {https://doi.org/10.48550/arXiv.2310.01783} {Can large language models provide useful feedback on research papers? {{A}} large-scale empirical analysis}.
\newblock \emph{Preprint}, arXiv:2310.01783.

\bibitem[{Lin(2004)}]{linROUGEPackageAutomatic2004}
Chin-Yew Lin. 2004.
\newblock \href {https://aclanthology.org/W04-1013/} {{{ROUGE}}: {{A Package}} for {{Automatic Evaluation}} of {{Summaries}}}.
\newblock In \emph{Text {{Summarization Branches Out}}}, pages 74--81, Barcelona, Spain. Association for Computational Linguistics.

\bibitem[{Liu et~al.(2024)Liu, Feng, Xue, Wang, Wu, Lu, Zhao, Deng, Zhang, Ruan et~al.}]{liu2024deepseek}
Aixin Liu, Bei Feng, Bing Xue, Bingxuan Wang, Bochao Wu, Chengda Lu, Chenggang Zhao, Chengqi Deng, Chenyu Zhang, Chong Ruan, et~al. 2024.
\newblock Deepseek-v3 technical report.
\newblock \emph{arXiv preprint arXiv:2412.19437}.

\bibitem[{Liu and Shah(2023)}]{liuReviewerGPTExploratoryStudy2023}
Ryan Liu and Nihar~B. Shah. 2023.
\newblock \href {https://doi.org/10.48550/arXiv.2306.00622} {{{ReviewerGPT}}? {{An Exploratory Study}} on {{Using Large Language Models}} for {{Paper Reviewing}}}.
\newblock \emph{Preprint}, arXiv:2306.00622.

\bibitem[{Qiao et~al.(2024)Qiao, Duan, Fang, Yang, Chen, Zhang, Wang, Lin, and Chen}]{qiaoPrismFrameworkDecoupling2024a}
Yuxuan Qiao, Haodong Duan, Xinyu Fang, Junming Yang, Lin Chen, Songyang Zhang, Jiaqi Wang, Dahua Lin, and Kai Chen. 2024.
\newblock \href {https://doi.org/10.48550/arXiv.2406.14544} {Prism: {{A Framework}} for {{Decoupling}} and {{Assessing}} the {{Capabilities}} of {{VLMs}}}.
\newblock \emph{Preprint}, arXiv:2406.14544.

\bibitem[{Qwen et~al.(2025)Qwen, Yang, Yang, Zhang, Hui, Zheng, Yu, Li, Liu, Huang, Wei, Lin, Yang, Tu, Zhang, Yang, Yang, Zhou, Lin, Dang, Lu, Bao, Yang, Yu, Li, Xue, Zhang, Zhu, Men, Lin, Li, Tang, Xia, Ren, Ren, Fan, Su, Zhang, Wan, Liu, Cui, Zhang, and Qiu}]{qwenQwen25TechnicalReport2025}
Qwen, An~Yang, Baosong Yang, Beichen Zhang, Binyuan Hui, Bo~Zheng, Bowen Yu, Chengyuan Li, Dayiheng Liu, Fei Huang, Haoran Wei, Huan Lin, Jian Yang, Jianhong Tu, Jianwei Zhang, Jianxin Yang, Jiaxi Yang, Jingren Zhou, Junyang Lin, Kai Dang, Keming Lu, Keqin Bao, Kexin Yang, Le~Yu, Mei Li, Mingfeng Xue, Pei Zhang, Qin Zhu, Rui Men, Runji Lin, Tianhao Li, Tianyi Tang, Tingyu Xia, Xingzhang Ren, Xuancheng Ren, Yang Fan, Yang Su, Yichang Zhang, Yu~Wan, Yuqiong Liu, Zeyu Cui, Zhenru Zhang, and Zihan Qiu. 2025.
\newblock \href {https://doi.org/10.48550/arXiv.2412.15115} {Qwen2.5 {{Technical Report}}}.
\newblock \emph{Preprint}, arXiv:2412.15115.

\bibitem[{Robertson(2023)}]{robertsonGPT4SlightlyHelpful2023}
Zachary Robertson. 2023.
\newblock \href {https://doi.org/10.48550/arXiv.2307.05492} {{{GPT4}} is {{Slightly Helpful}} for {{Peer-Review Assistance}}: {{A Pilot Study}}}.
\newblock \emph{Preprint}, arXiv:2307.05492.

\bibitem[{Shen et~al.(2022)Shen, Cheng, Zhou, Bing, You, and Si}]{shenMReDMetaReviewDataset2022}
Chenhui Shen, Liying Cheng, Ran Zhou, Lidong Bing, Yang You, and Luo Si. 2022.
\newblock \href {https://doi.org/10.18653/v1/2022.findings-acl.198} {{{MReD}}: {{A Meta-Review Dataset}} for {{Structure-Controllable Text Generation}}}.
\newblock In \emph{Findings of the {{Association}} for {{Computational Linguistics}}: {{ACL}} 2022}, pages 2521--2535, Dublin, Ireland. Association for Computational Linguistics.

\bibitem[{Tan et~al.(2024)Tan, Lyu, Li, Gao, Wei, Ma, Liu, and Li}]{tanPeerReviewMultiTurn2024}
Cheng Tan, Dongxin Lyu, Siyuan Li, Zhangyang Gao, Jingxuan Wei, Siqi Ma, Zicheng Liu, and Stan~Z. Li. 2024.
\newblock \href {https://doi.org/10.48550/arXiv.2406.05688} {Peer {{Review}} as {{A Multi-Turn}} and {{Long-Context Dialogue}} with {{Role-Based Interactions}}}.
\newblock \emph{Preprint}, arXiv:2406.05688.

\bibitem[{Wang and Wan(2018)}]{wangSentimentAnalysisPeer2018}
Ke~Wang and Xiaojun Wan. 2018.
\newblock \href {https://doi.org/10.1145/3209978.3210056} {Sentiment {{Analysis}} of {{Peer Review Texts}} for {{Scholarly Papers}}}.
\newblock In \emph{The 41st {{International ACM SIGIR Conference}} on {{Research}} \& {{Development}} in {{Information Retrieval}}}, {{SIGIR}} '18, pages 175--184, New York, NY, USA. Association for Computing Machinery.

\bibitem[{Wei et~al.(2022)Wei, Wang, Schuurmans, Bosma, Ichter, Xia, Chi, Le, and Zhou}]{weiChainofThoughtPromptingElicits2022}
Jason Wei, Xuezhi Wang, Dale Schuurmans, Maarten Bosma, Brian Ichter, Fei Xia, Ed~Chi, Quoc~V. Le, and Denny Zhou. 2022.
\newblock \href {https://papers.neurips.cc/paper\_files/paper/2022/hash/9d5609613524ecf4f15af0f7b31abca4-Abstract-Conference.html} {Chain-of-{{Thought Prompting Elicits Reasoning}} in {{Large Language Models}}}.
\newblock \emph{Advances in Neural Information Processing Systems}, 35:24824--24837.

\bibitem[{Xu et~al.(2025)Xu, Jin, Li, Song, Sun, and Yuan}]{xuLLaVACoTLetVision2025}
Guowei Xu, Peng Jin, Hao Li, Yibing Song, Lichao Sun, and Li~Yuan. 2025.
\newblock \href {https://doi.org/10.48550/arXiv.2411.10440} {{{LLaVA-CoT}}: {{Let Vision Language Models Reason Step-by-Step}}}.
\newblock \emph{Preprint}, arXiv:2411.10440.

\bibitem[{Yuan et~al.(2021)Yuan, Liu, and Neubig}]{yuanCanWeAutomate2021}
Weizhe Yuan, Pengfei Liu, and Graham Neubig. 2021.
\newblock \href {https://doi.org/10.48550/arXiv.2102.00176} {Can {{We Automate Scientific Reviewing}}?}
\newblock \emph{Preprint}, arXiv:2102.00176.

\bibitem[{Zhang et~al.(2024)Zhang, Xu, Zhang, Liu, Hooi, and Deng}]{zhangExploringCollaborationMechanisms2024}
Jintian Zhang, Xin Xu, Ningyu Zhang, Ruibo Liu, Bryan Hooi, and Shumin Deng. 2024.
\newblock \href {https://doi.org/10.18653/v1/2024.acl-long.782} {Exploring {{Collaboration Mechanisms}} for {{LLM Agents}}: {{A Social Psychology View}}}.
\newblock In \emph{Proceedings of the 62nd {{Annual Meeting}} of the {{Association}} for {{Computational Linguistics}} ({{Volume}} 1: {{Long Papers}})}, pages 14544--14607, Bangkok, Thailand. Association for Computational Linguistics.

\bibitem[{Zhao et~al.(2024)Zhao, Xing, Dou, Tian, Tai, Yang, Cheng, and Li}]{zhaoWordsWorthNewborn2024}
Penghai Zhao, Qinghua Xing, Kairan Dou, Jinyu Tian, Ying Tai, Jian Yang, Ming-Ming Cheng, and Xiang Li. 2024.
\newblock \href {https://doi.org/10.48550/arXiv.2408.03934} {From {{Words}} to {{Worth}}: {{Newborn Article Impact Prediction}} with {{LLM}}}.
\newblock \emph{Preprint}, arXiv:2408.03934.

\bibitem[{Zheng et~al.(2024)Zheng, Zhang, Zhang, Ye, Luo, Feng, and Ma}]{zhengLlamaFactoryUnifiedEfficient2024}
Yaowei Zheng, Richong Zhang, Junhao Zhang, Yanhan Ye, Zheyan Luo, Zhangchi Feng, and Yongqiang Ma. 2024.
\newblock \href {https://doi.org/10.48550/arXiv.2403.13372} {{{LlamaFactory}}: {{Unified Efficient Fine-Tuning}} of 100+ {{Language Models}}}.
\newblock \emph{Preprint}, arXiv:2403.13372.

\bibitem[{Zhong et~al.(2024)Zhong, Liu, Pan, Zhang, Zhou, Liang, Wu, Lyu, Shu, Yu, Cao, Jiang, Chen, Li, Chen, Hu, Liu, Zhao, Xu, Dai, Zhao, Zhang, Zhao, Yang, Chen, Wang, Ruan, Wang, Zhao, Zhang, Ren, Qin, Chen, Li, Zidan, Jahin, Chen, Xia, Holmes, Zhuang, Wang, Xu, Xia, Yu, Tang, Yang, Sun, Yang, Lu, Wang, Chai, Li, Lu, Sun, Zhang, Ge, Hu, Zhang, Zhou, Zhang, Zhang, Liu, Jiang, Kong, Xiang, Ren, Liu, Jiang, Bao, Zhang, Li, Li, Liu, Shen, Sikora, Zhai, Zhu, and Liu}]{zhongEvaluationOpenAIO12024}
Tianyang Zhong, Zhengliang Liu, Yi~Pan, Yutong Zhang, Yifan Zhou, Shizhe Liang, Zihao Wu, Yanjun Lyu, Peng Shu, Xiaowei Yu, Chao Cao, Hanqi Jiang, Hanxu Chen, Yiwei Li, Junhao Chen, Huawen Hu, Yihen Liu, Huaqin Zhao, Shaochen Xu, Haixing Dai, Lin Zhao, Ruidong Zhang, Wei Zhao, Zhenyuan Yang, Jingyuan Chen, Peilong Wang, Wei Ruan, Hui Wang, Huan Zhao, Jing Zhang, Yiming Ren, Shihuan Qin, Tong Chen, Jiaxi Li, Arif~Hassan Zidan, Afrar Jahin, Minheng Chen, Sichen Xia, Jason Holmes, Yan Zhuang, Jiaqi Wang, Bochen Xu, Weiran Xia, Jichao Yu, Kaibo Tang, Yaxuan Yang, Bolun Sun, Tao Yang, Guoyu Lu, Xianqiao Wang, Lilong Chai, He~Li, Jin Lu, Lichao Sun, Xin Zhang, Bao Ge, Xintao Hu, Lian Zhang, Hua Zhou, Lu~Zhang, Shu Zhang, Ninghao Liu, Bei Jiang, Linglong Kong, Zhen Xiang, Yudan Ren, Jun Liu, Xi~Jiang, Yu~Bao, Wei Zhang, Xiang Li, Gang Li, Wei Liu, Dinggang Shen, Andrea Sikora, Xiaoming Zhai, Dajiang Zhu, and Tianming Liu. 2024.
\newblock \href {https://doi.org/10.48550/arXiv.2409.18486} {Evaluation of {{OpenAI}} o1: {{Opportunities}} and {{Challenges}} of {{AGI}}}.
\newblock \emph{Preprint}, arXiv:2409.18486.

\bibitem[{Zhou et~al.(2024)Zhou, Chen, and Yu}]{zhouLLMReliableReviewer2024}
Ruiyang Zhou, Lu~Chen, and Kai Yu. 2024.
\newblock \href {https://aclanthology.org/2024.lrec-main.816/} {Is {{LLM}} a {{Reliable Reviewer}}? {{A Comprehensive Evaluation}} of {{LLM}} on {{Automatic Paper Reviewing Tasks}}}.
\newblock In \emph{Proceedings of the 2024 {{Joint International Conference}} on {{Computational Linguistics}}, {{Language Resources}} and {{Evaluation}} ({{LREC-COLING}} 2024)}, pages 9340--9351, Torino, Italia. {ELRA and ICCL}.

\bibitem[{Zhu et~al.(2024)Zhu, Qu, Dong, Ruan, Tong, He, and Cheng}]{zhu2024llama}
Tong Zhu, Xiaoye Qu, Daize Dong, Jiacheng Ruan, Jingqi Tong, Conghui He, and Yu~Cheng. 2024.
\newblock Llama-moe: Building mixture-of-experts from llama with continual pre-training.
\newblock In \emph{Proceedings of the 2024 Conference on Empirical Methods in Natural Language Processing}, pages 15913--15923.

\bibitem[{Zhu et~al.(2018)Zhu, Lu, Zheng, Guo, Zhang, Wang, and Yu}]{zhuTexygenBenchmarkingPlatform2018}
Yaoming Zhu, Sidi Lu, Lei Zheng, Jiaxian Guo, Weinan Zhang, Jun Wang, and Yong Yu. 2018.
\newblock \href {https://doi.org/10.48550/arXiv.1802.01886} {Texygen: {{A Benchmarking Platform}} for {{Text Generation Models}}}.
\newblock \emph{Preprint}, arXiv:1802.01886.

\bibitem[{Zhuang et~al.(2025)Zhuang, Chen, Xu, Jiang, and Lin}]{zhuangLargeLanguageModels2025}
Zhenzhen Zhuang, Jiandong Chen, Hongfeng Xu, Yuwen Jiang, and Jialiang Lin. 2025.
\newblock \href {https://doi.org/10.48550/arXiv.2501.10326} {Large language models for automated scholarly paper review: {{A}} survey}.
\newblock \emph{Preprint}, arXiv:2501.10326.

\end{thebibliography}
\clearpage
\appendix
\section{More Details on ReviewBench and Evaluation Metrics}
\label{appendix:metrics}

\subsection{Research Areas in ReviewBench}
\begin{table}
\centering
\small
\begin{tabular}{cc}
\hline
Research Domain                    &       \# of papers    \\\hline
Machine Learning and Deep Learning & 25              \\
NLP and Generative Models          & 22              \\
Reinforcement Learning             & 16              \\
Privacy and Security               & 14              \\
Computer Vision                    & 9               \\
Graph Neural Networks              & 6               \\
Mathematics and Optimization       & 5               \\
Application                        & 3               \\ \hline
\end{tabular}%

\caption{The research domains and the number of the papers in ReviewBench}
\label{table:area}
\label{tab:my-table}
\end{table}

Table \ref{table:area} presents the research domains of the papers in ReviewBench and their corresponding quantities.

\subsection{More Details on Evaluation Metrics}
\subsubsection*{Distinct}
Distinct measures the diversity of the generated text by calculating the proportion of unique n-grams in the generated text. A higher Distinct score indicates more variety.

\[
\text{Distinct}_n = \frac{\text{\# of unique n-grams}}{\text{Total \# of n-grams}}
\]

\subsubsection*{Self-BLEU}
Self-BLEU measures the diversity of the generated text by computing the BLEU score between the generated text and its multiple variants. A higher Self-BLEU score indicates less diversity.

\[
\text{Self-BLEU} = \frac{1}{N} \sum_{i=1}^{N} \text{BLEU}(r_i, \hat{r}_i)
\]
Where:
\begin{itemize}
  \item \( r_i \) is the i-th variant of the generated text.
  \item \( \hat{r}_i \) is another variant with which the i-th variant is compared.
  \item \( N \) is the total number of generated texts.
\end{itemize}

In order to compute the overall score, we employ \textbf{Inverse Self-BLEU} in this paper, where a higher score indicates greater diversity.

\subsubsection*{ROUGE-1}
ROUGE-1 measures the overlap of unigrams (single words) between the generated text and reference text.

\[
\text{Precision} = \frac{\text{\# of matching unigrams}}{\text{Total \# of unigrams in generated text}}
\]

\[
\text{Recall} = \frac{\text{\# of matching unigrams}}{\text{Total \# of unigrams in reference text}}
\]

\[
\text{F1-score} = 2 \times \frac{\text{Precision} \times \text{Recall}}{\text{Precision} + \text{Recall}}
\]

\subsubsection*{ROUGE-L}
ROUGE-L measures the overlap of the longest common subsequence (LCS) between the generated text and the reference text.

\[
\text{Precision} = \frac{\text{LCS length}}{\text{Length of generated text}}
\]

\[
\text{Recall} = \frac{\text{LCS length}}{\text{Length of reference text}}
\]

\[
\text{F1-score} = 2 \times \frac{\text{Precision} \times \text{Recall}}{\text{Precision} + \text{Recall}}
\]

In our work, the calculation of the ROUGE series metrics is done using the rouge\_score library in Python.

\section{Implementation Details}
\label{appendix:implement}
Our framework utilizes the Llama-3.1-8B-Instruct model trained on NVIDIA A100 GPUs. When generating review comments, we set the number of reviewers, $N$, to 3. The judge for the Review Arena is GPT-4o. We selected advanced open-source models
such as Llama-3.1 \cite{dubey2024llama}, Qwen2.5 \cite{qwenQwen25TechnicalReport2025}, 
and closed-source models
including GPT-4o, Claude-3.5, and Deepseek\cite{liu2024deepseek,guo2025deepseek} 
for comparison. We performed supervised fine-tuning on the base model Llama-3.1-8B-Instruct implemented using Llama-Factory \cite{zhengLlamaFactoryUnifiedEfficient2024}, setting the number of training epochs to 3, gradient accumulation steps to 32, and warm-up steps to 100. The learning rate was set to 1e-5, with training acceleration implemented using liger-kernel and memory scheduling managed through the Zero3 strategy. The total training cost was approximately 1000 GPU hours.

\section{Human Evaluation Questionnaire}
Table \ref{tab:human-eval} presents the questionnaire used for human evaluation, detailing the meaning of each evaluation dimension and instructing evaluators to rate the LLM-generated review comments on a scale from 1 to 5. Human-written reviews were also provided alongside as a reference for assessing the relevance and alignment of the generated reviews.

\label{questionnaire}
\begin{table*}[!t]
\centering
\small
\begin{tabular}{p{2cm}|p{7.3cm}|c}
\hline
Dimension &
  Description \centering &
  Scoring Guide \\ \hline
Soundness \& Validity \centering&
  Does the review provide reasonable, objective evaluations and suggestions based on the content of the paper? &
  \begin{tabular}[c]{@{}c@{}}1 = Groundless or incorrect judgments;\\ 5 = Rigorous and well-supported analysis\end{tabular} \\
Clarity \& Organization \centering&
  Is the review well-structured, logically organized, and easy to understand? &
  \begin{tabular}[c]{@{}c@{}}1 = Disorganized and unclear; \\ 5 = Well-structured and clearly articulated\end{tabular} \\
Constructiveness \centering&
  Does the review offer concrete and valuable suggestions for improvement, rather than just identifying problems? &
  \begin{tabular}[c]{@{}c@{}}1 = Vague or unhelpful feedback; \\ 5 = Clear, specific, and helpful feedback\end{tabular} \\
Consistency with Human Review \centering&
  How consistent is the review with real human expert reviews in terms of conclusions, focus points, and suggestions? &
  \begin{tabular}[c]{@{}c@{}}1 = Significantly inconsistent; \\ 5 = Highly consistent\end{tabular} \\
Position Clarity \centering&
  Does the review clearly state a stance? &
  \begin{tabular}[c]{@{}c@{}}1 = No clear stance or vague position; \\ 5 = Clear and reasonable stance\end{tabular} \\ \hline
\end{tabular}
\caption{The questionnaire used for human evaluation}
\label{tab:human-eval}
\end{table*}

\section{Prompt Templates}
\label{appendix:prompt}

\begin{figure}[!t]
    \centering
    \includegraphics[width=\linewidth]{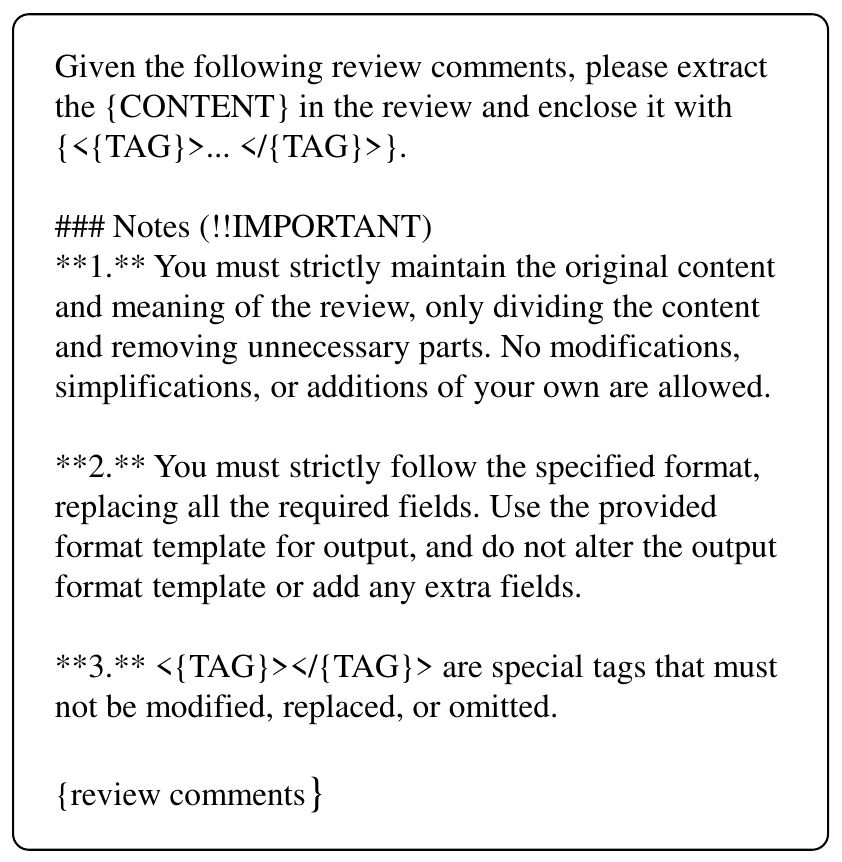}
    \caption{Prompt for transcription}
    \label{fig:prompt-transcription}
\end{figure}

\begin{figure}[!t]
    \centering
    \includegraphics[width=\linewidth]{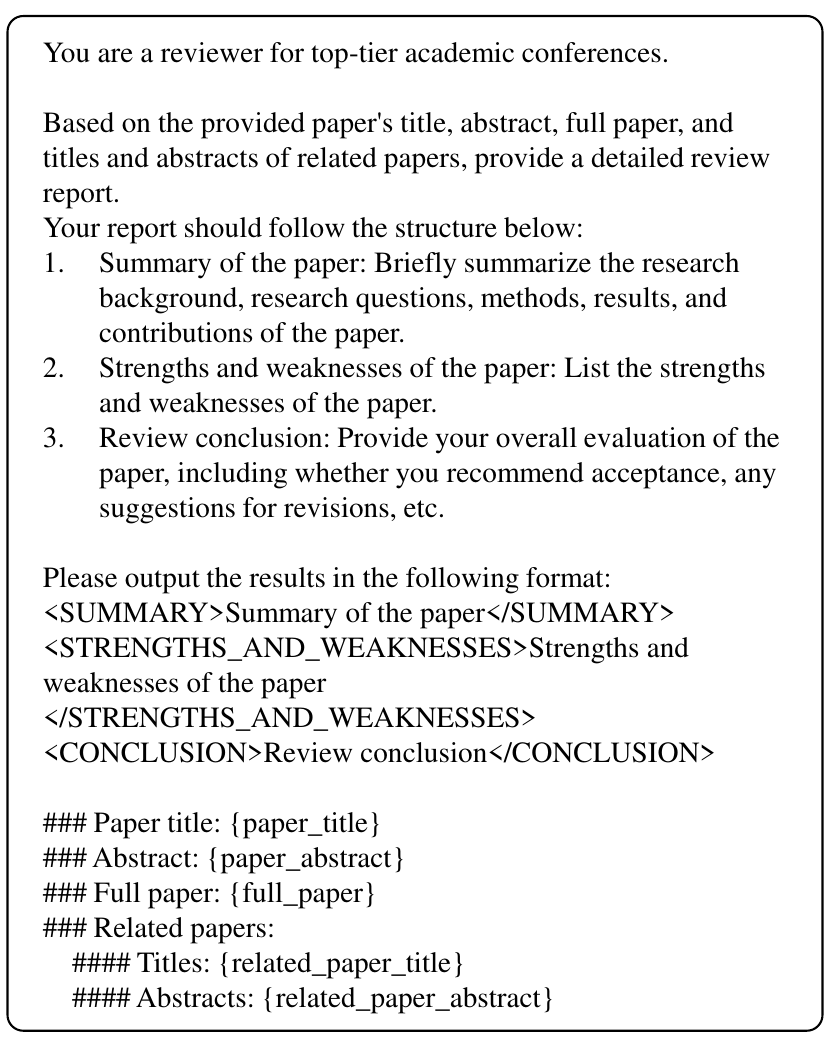}
    \caption{Prompt for review generation for Review Agents}
    \label{fig:prompt-review}
\end{figure}

\begin{figure}[!t]
    \centering
    \includegraphics[width=\linewidth]{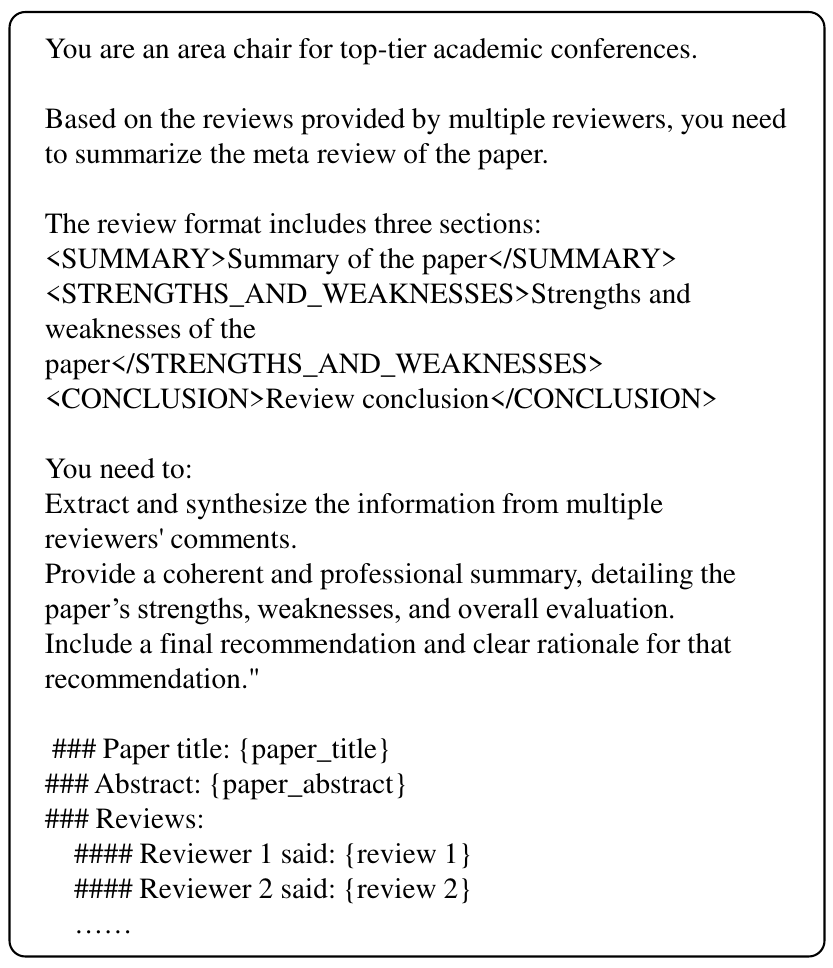}
    \caption{Prompt for meta review generation}
    \label{fig:prompt-meta}
\end{figure}

\begin{figure}[!t]
    \centering
    \includegraphics[width=\linewidth]{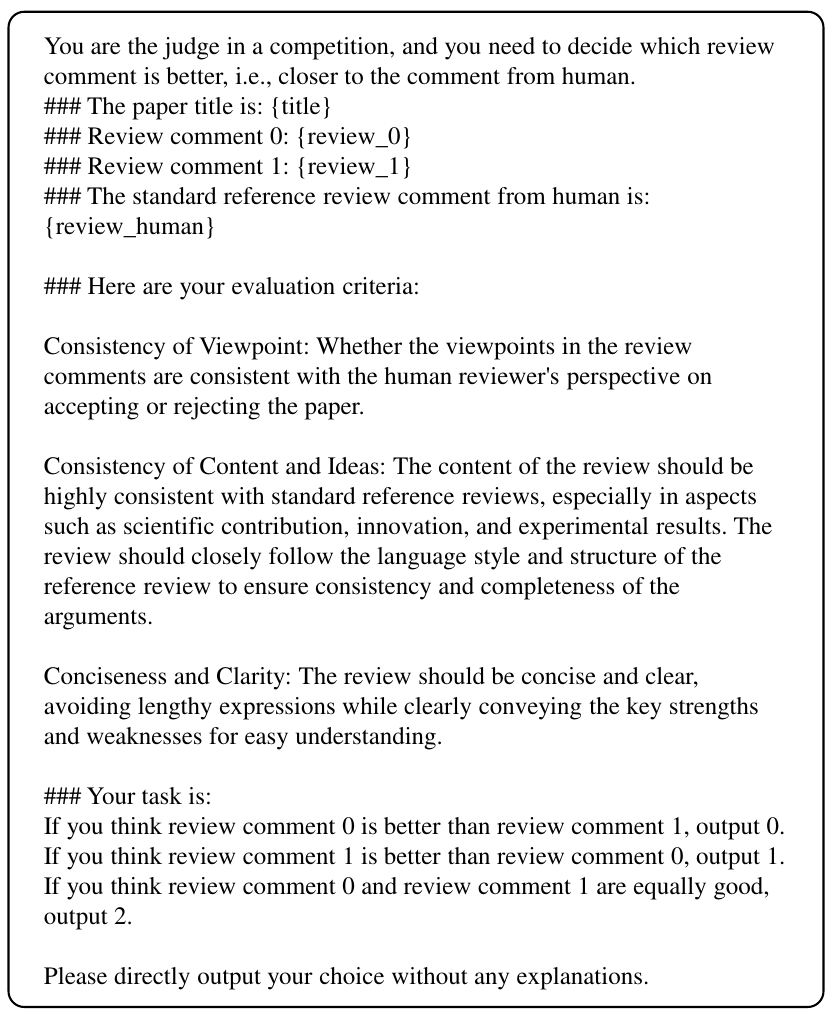}
    \caption{Prompt used in Review Arena}
    \label{fig:prompt-arena}
\end{figure}

\begin{itemize}
    \item Transcription: Figure \ref{fig:prompt-transcription}
    \item Review generation: Figure \ref{fig:prompt-review}
    \item Meta review generation: Figure \ref{fig:prompt-meta}
    \item Review Arena: Figure \ref{fig:prompt-arena}
\end{itemize}

\section{A Case Study of a Complete Process}
\label{appendix:case}

\begin{figure*}[!t]
    \centering
    \includegraphics[width=\linewidth]{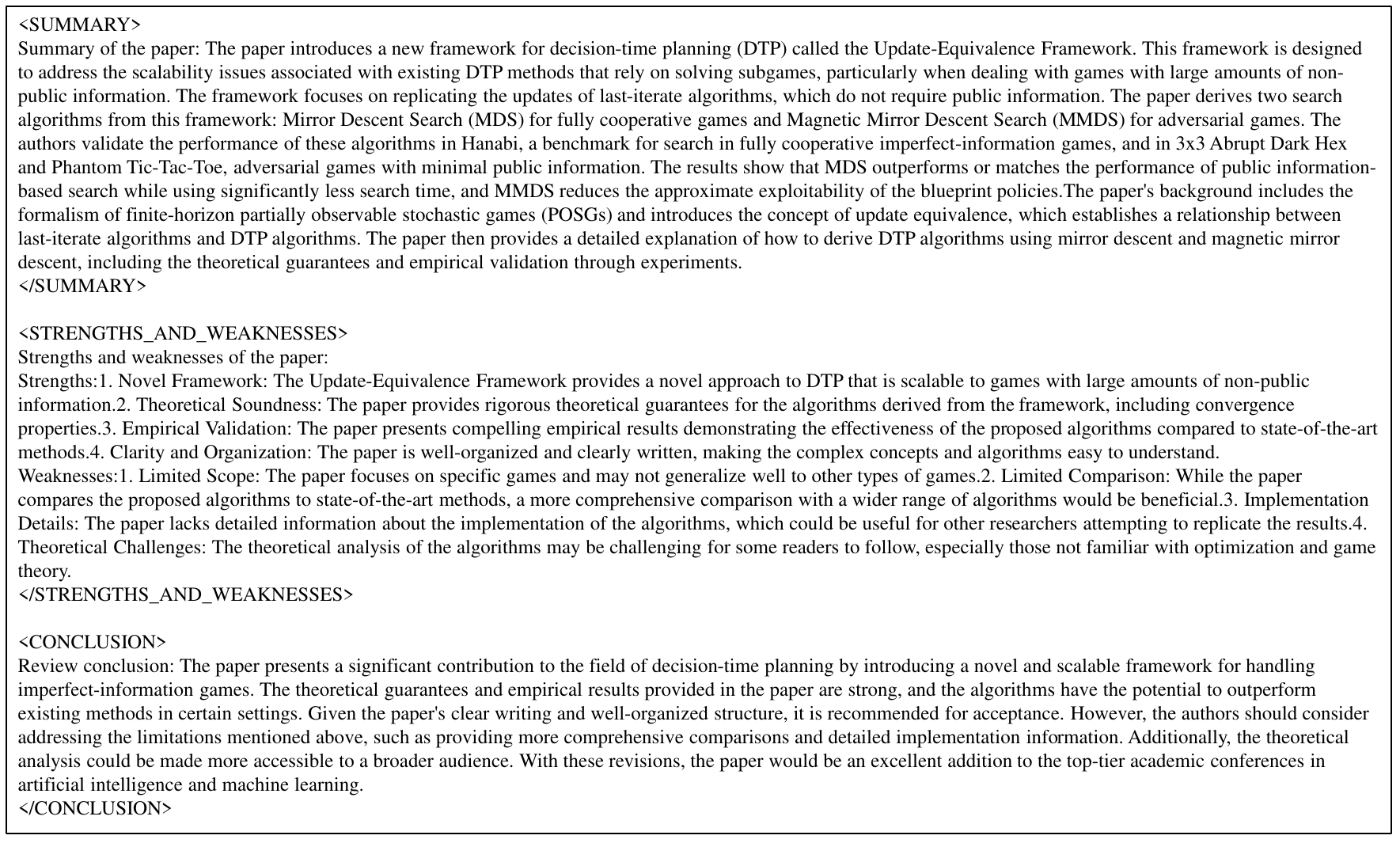}
    \caption{Review from reviewer agent 1}
    \label{fig:case_review1}
\end{figure*}

\begin{figure*}[!t]
    \centering
    \includegraphics[width=\linewidth]{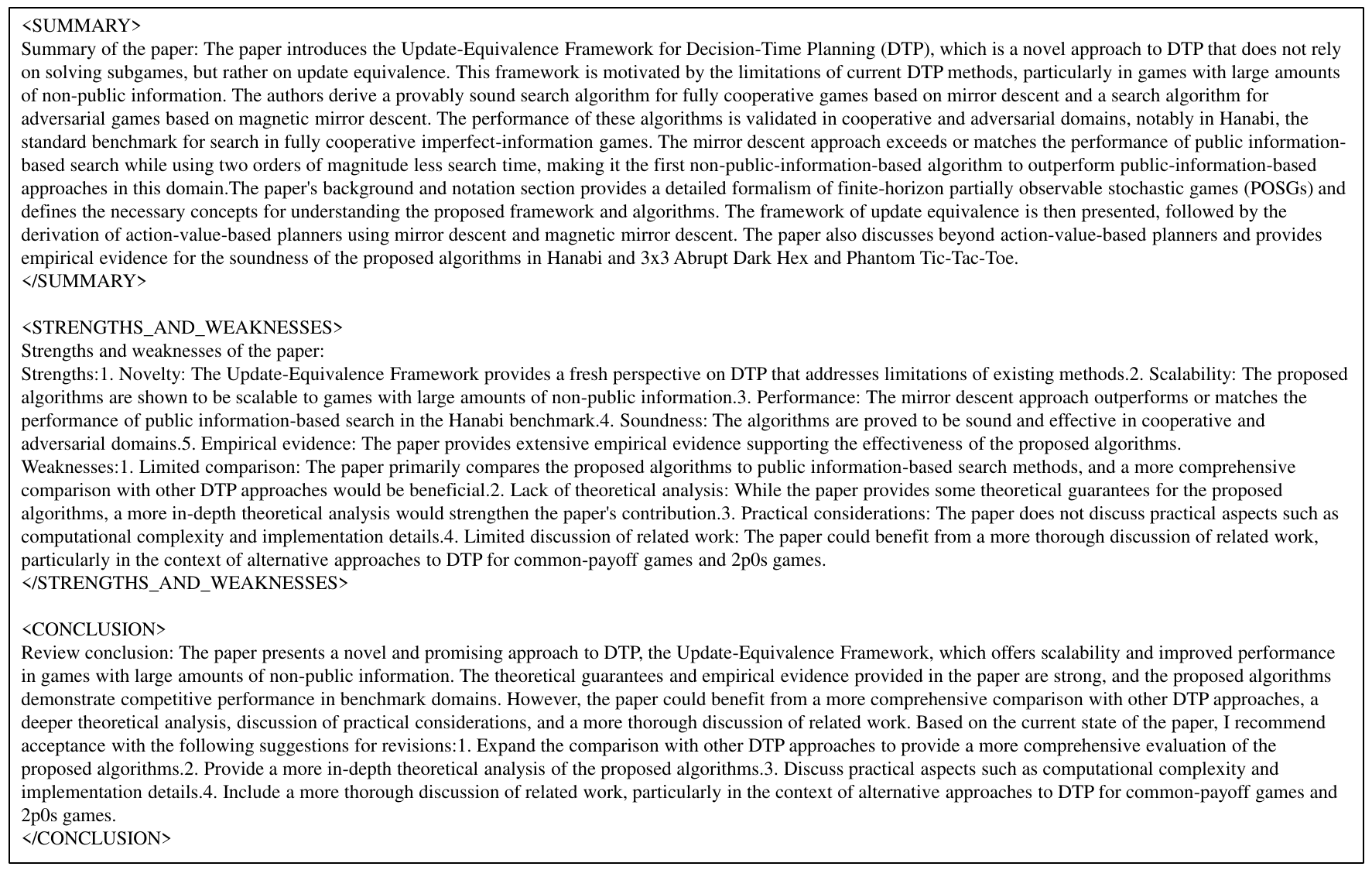}
    \caption{Review from reviewer agent 2}
    \label{fig:case_review2}
\end{figure*}

\begin{figure*}[!t]
    \centering
    \includegraphics[width=\linewidth]{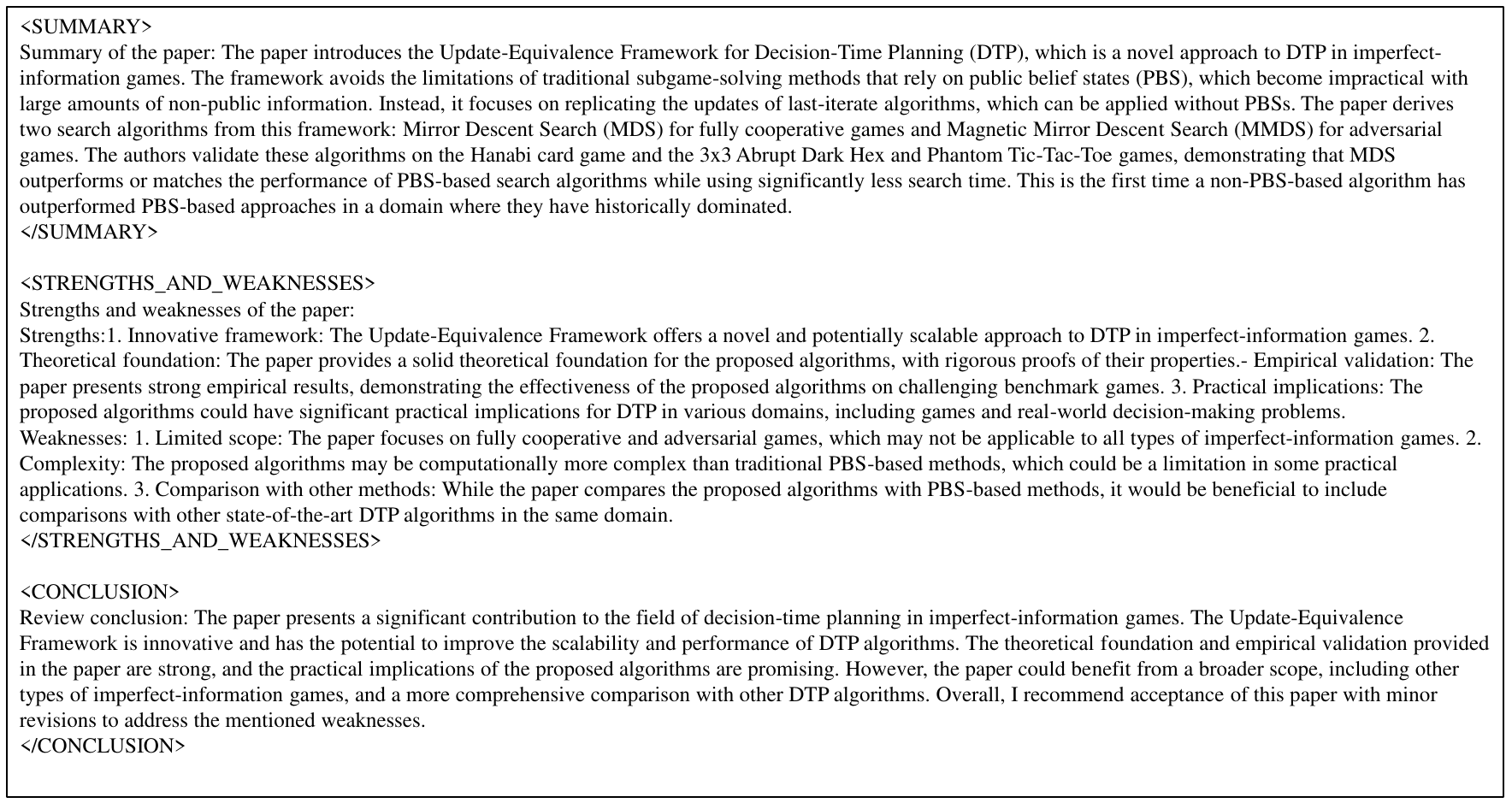}
    \caption{Review from reviewer agent 3}
    \label{fig:case_review3}
\end{figure*}

\begin{figure*}[!t]
    \centering
    \includegraphics[width=\linewidth]{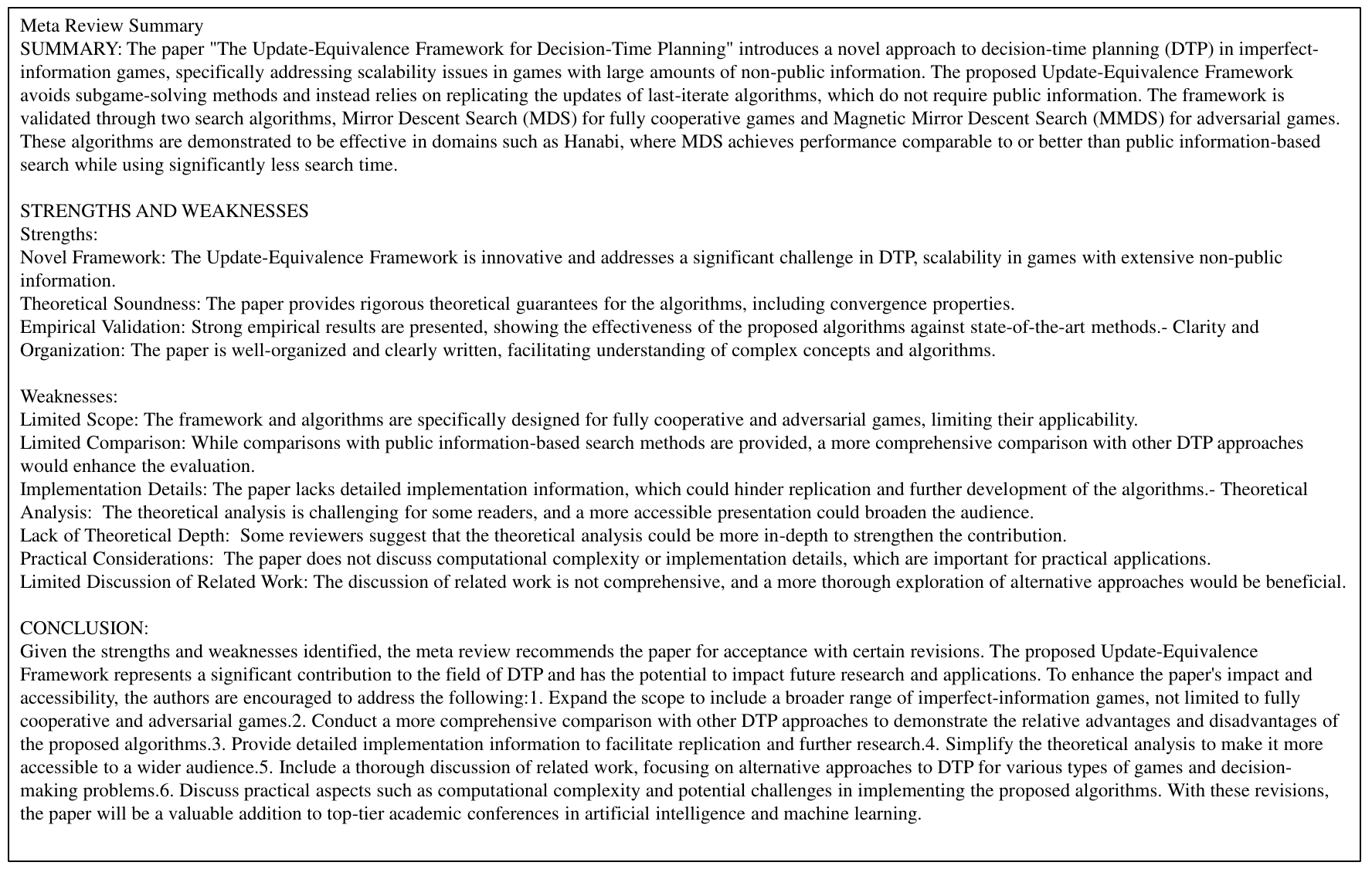}
    \caption{Meta review from area chair agent}
    \label{fig:case_review_meta}
\end{figure*}

\begin{figure*}[!t]
    \centering
    \includegraphics[width=\linewidth]{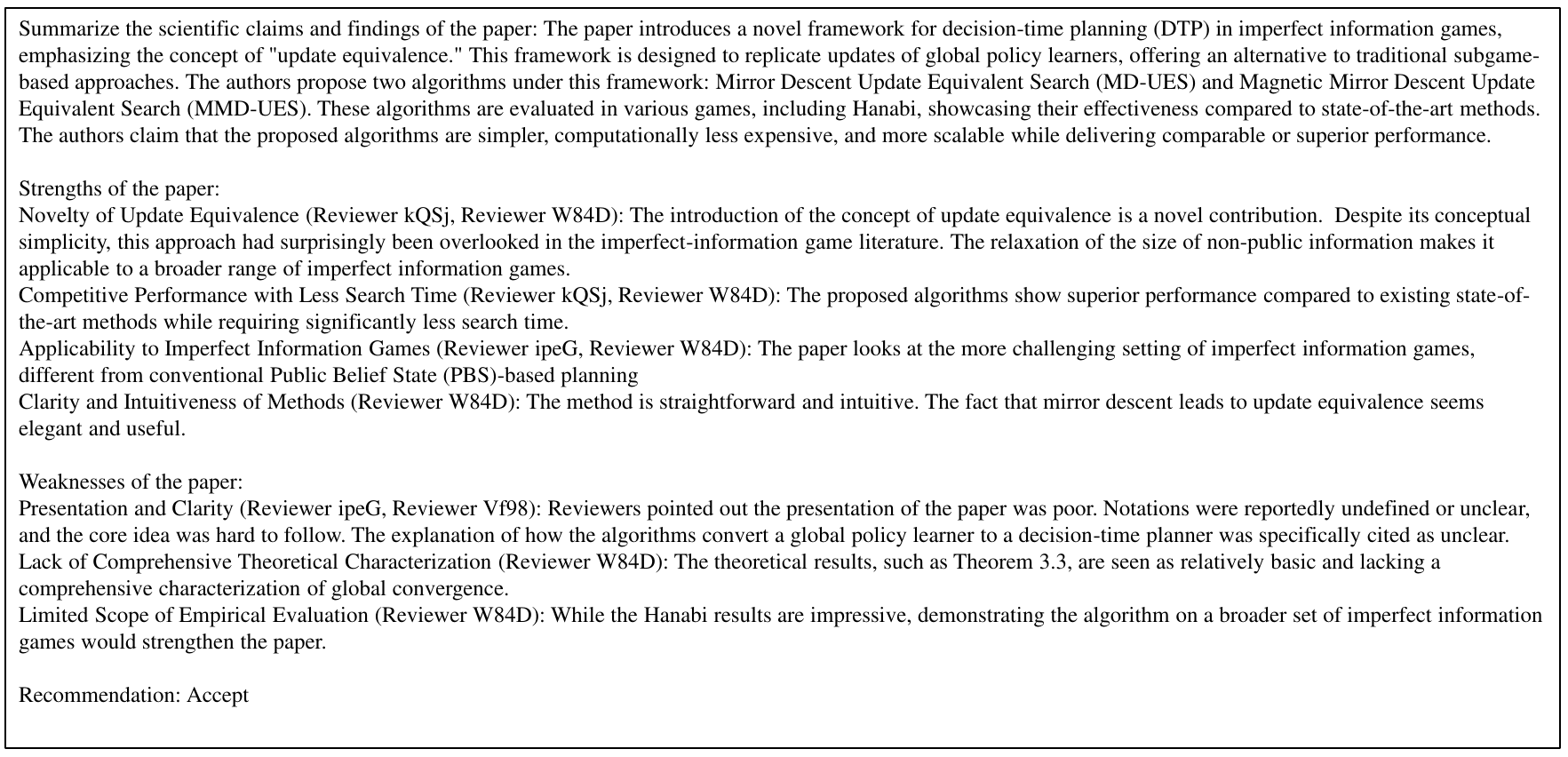}
    \caption{Meta review from human}
    \label{fig:case_review_meta_human}
\end{figure*}

In Figures \ref{fig:case_review1} to \ref{fig:case_review_meta}, we present the output of a complete review process, providing genuine human review comments for reference (Figure \ref{fig:case_review_meta_human}). The results indicate that the area chair agent in the ReviewAgents framework effectively synthesizes feedback from different reviewers and generates a well-rounded conclusion. The review conclusions produced by the agent show a high degree of overlap with those made by human reviewers, offering similar feedback on aspects such as the novelty, clarity, lack of theoretical depth, and limited scope of the paper. The final recommendation, similarly, results in an "accept" decision.

\end{document}